\documentclass{article}

\usepackage[preprint]{corl_2026} 

\usepackage[table]{xcolor}
\definecolor{ourblue}{HTML}{E6F2FB}
\usepackage{amsmath,amssymb}
\usepackage{graphicx}
\usepackage{capt-of}
\usepackage{booktabs}
\usepackage{multirow}
\usepackage{subcaption}
\usepackage{wrapfig}
\usepackage{makecell}
\usepackage{tikz}

\usetikzlibrary{positioning,arrows.meta,calc,shapes.geometric,fit,backgrounds}

\title{Enabling Robust Cloth Manipulation via Inference-Time Simulator-in-the-Loop Refinement}

\author{
  Xin Liu$^{1,2,*}$,
  Yulin Li$^{1,*}$,
  Ziming Li$^{1}$,
  Pengyu Jing$^{1}$,
  Zhenhao Huang$^{1}$, \\
  \bfseries
  Bingyang Zhou$^{1}$,
  Ziqiu Zeng$^{1}$,
  Siyuan Luo$^{1}$,
  Chenkun Qi$^{2}$,
  Fan Shi$^{1,\dagger}$ \\[0.5em]
  $^{1}$ National University of Singapore,
  $^{2}$ Shanghai Jiao Tong University \\ [0.3em]
  $^{*}$ Equal contribution. \quad $^{\dagger}$ Corresponding author.
}

\begin{document}
\maketitle

\begin{abstract}
\textit{Simulator-in-the-loop} optimization offers a promising inference-time mechanism for robot manipulation. It uses a physical simulator as a backend rollout engine to evaluate candidate trajectories in parallel and refine nominal actions online, a paradigm shown to be effective in rigid-body manipulation where state and contact are relatively tractable. We bring this paradigm to real-world cloth manipulation from a single RGB input through three pillars. \emph{(i)} We design a scalable synthetic-data generation and inference-time rollout pipeline built on \textsc{FLASH}, a deformable-object simulator that provides a practical balance among physical fidelity, numerical stability, and rollout efficiency. \emph{(ii)} We develop a real-to-sim module, trained purely on synthetic data, that maps a single RGB observation to simulation-compatible cloth state by fusing pretrained visual features with learnable canonical tokens. \emph{(iii)} We perform online planning by coupling a sparse-mesh rollout backend with prior-guided MPPI, anchored at an offline-distilled policy trajectory, preserving manipulation-relevant deformation and contact while enabling sufficient parallel rollout batches. Real-robot experiments show higher success rates than baseline methods and closed-loop correction under mid-fold perturbations. Project page: \url{https://silr-cloth.github.io/}
\end{abstract}
\keywords{Deformable Object Manipulation, Simulator-in-the-Loop Optimization, Cloth State Estimation}

\section{Introduction}
\label{sec:intro}
Deformable-object manipulation presents fundamental challenges: infinite-dimensional configuration spaces, continuously evolving local geometry, severe self-occlusion, and pervasive frictional contact. These factors substantially complicate both state estimation and action reasoning, making reliable cloth manipulation considerably more challenging~\cite{zhu2022challenges, arriolarios2020modeling, Xingyu2020SoftGym, hu2025real}.

Imitation learning has become a dominant paradigm for deformable-object manipulation, bypassing explicit dynamics modeling by fitting policies to the distribution of expert demonstrations and robot interaction data~\cite{chi2023diffusion, fu2024mobilealoha, ze2024dp3, zhao2023learning}. Recent robot foundation models~\cite{zitkovich2023rt2, kim2025openvla,yang2024unisim, maes2026leworldmodel, assran2025v} further scale this idea and show promising task-level generalization~\cite{black2024pi_0, padalkar2024open}. However, at the deployment stage, these policies have limited ability to correct their own execution errors. Small deviations in perception, grasp selection, timing, or action prediction can accumulate through contact-rich cloth dynamics and drive the system away from the intended task state~\cite{pmlr-v15-ross11a, spencer2021feedback}. Reliable cloth manipulation, therefore, requires an inference-time mechanism that refines policy proposals and recovers from accumulated errors.

High-fidelity deformable-object simulators offer a promising foundation for physically grounded inference-time reasoning in cloth manipulation. Recent work has explored this direction by using simulators to generate data for diffusion-based neural dynamics models, paired with full-state reconstruction from partially observed point clouds~\cite{pmlr-v305-tian25c}. We argue instead that the physical simulator itself is the most natural forward dynamics model~\cite{howell2022mujocompc}: direct simulator rollout avoids task- or asset-specific dynamics training, removes the additional approximation error of learned predictors, and can serve as a general physical reasoning engine across deformable assets.

This raises the central question of this work: \textit{Can we directly use deformable-object simulators as inference-time forward models for real-world cloth manipulation?} Answering this question requires solving the full real-to-sim-to-real loop. The simulator must be physically faithful and fast enough for online rollout. The real cloth state must be recovered accurately enough to synchronize a simulated twin. The rollout representation must balance state granularity, dynamics fidelity, and parallel candidate diversity under a limited control budget. If any part fails, the optimized action becomes misaligned with the physical scene or too expensive to compute online.

\textbf{Contributions.}
We structure the investigation around three coupled system-level questions: \emph{(i)} how simulator fidelity, numerical stability, and rollout efficiency affect inference-time refinement for cloth manipulation; \emph{(ii)} how to recover simulation-compatible cloth states from practical real-world observations; and \emph{(iii)} how to refine nominal policy trajectories through closed-loop inference-time physics rollouts under limited online budgets.

\begin{figure}[t]
    \centering
\includegraphics[width=\textwidth]{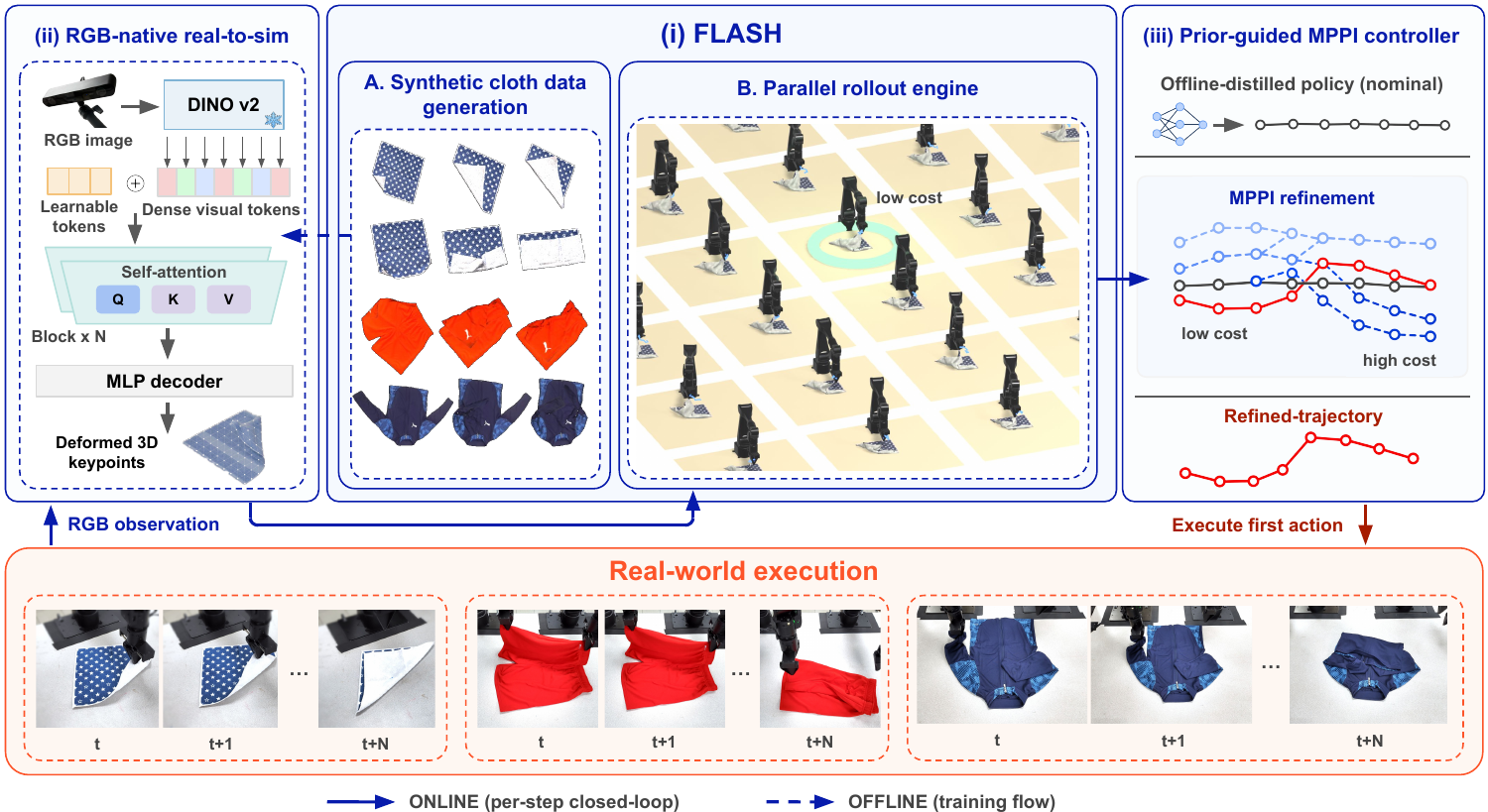}
\caption{
\textbf{Overview of the proposed framework.} Offline, we use FLASH to generate synthetic data for real-to-sim training. Online, RGB observations initialize physics rollouts that refine the prior policy through MPPI for closed-loop hardware execution.
}
\vspace{-0.5cm}
\label{fig:pipeline}
\end{figure}

Based on this joint study, we propose a unified \textit{simulator-in-the-loop} framework for real-world cloth manipulation shown in Fig.~\ref{fig:pipeline}. The framework contains two coupled flows. Offline, we use \textsc{FLASH}~\cite{siyuan2026FALSH} to generate scalable synthetic cloth-deformation data and train an RGB-native real-to-sim module with pretrained visual features, canonical query tokens, and random grasping-point masking. Online, a single RGB observation is lifted into a sparse simulation-compatible cloth state, which synchronizes a sparse cloth-mesh rollout backend. A prior-guided MPPI controller then refines the nominal policy trajectory through parallel physics rollouts and applies the optimized action in a receding-horizon loop on hardware. The proposed pipeline provides a general mechanism for refining prior policies through closed-loop online physics inference under a limited rollout budget. To our knowledge, this is the first framework that directly uses a deformable-object simulator as an inference-time forward model for cloth manipulation within an online receding-horizon control loop. Real-robot experiments show that replacing or weakening any single component degrades closed-loop performance, and that the full system achieves higher success rates and closed-loop correction under mid-fold perturbations, with further evaluations demonstrating generalization potential across deformable assets of different shapes and local deformation patterns.


\section{Related Works}
\label{sec:related}
We use \emph{simulator-in-the-loop control} to refer to methods that refine policy execution with an inference-time forward model. Given an estimated state and candidate actions, the model predicts short-horizon physical consequences for online optimization, and the selected action is applied in a receding-horizon manner.
We review prior work along three axes central to this paradigm.

\textbf{State Estimation.}
Cloth state estimation must recover a high-dimensional, continuously deforming and self-occluded configuration, rather than the compact 6D pose that suffices for rigid objects~\cite{Xiang-RSS-18, wen2023bundlesdf, wen2024foundationpose}. Prior work reconstructs garment or cloth geometry from partial observations, including garment completion from point clouds~\cite{9710790}, mesh reconstruction with occlusion reasoning~\cite{Huang-RSS-22}, template-based reconstruction of crumpled cloths from top-view depth~\cite{wang2024trtm}, diffusion-based mesh prediction from multi-view partial point clouds~\cite{pmlr-v305-tian25c}, and RGB-based refinement through 3D Gaussian splatting~\cite{pmlr-v270-longhini25a}. Other methods avoid explicit full-state reconstruction by feeding point clouds~\cite{chen2025metafold} or 3D keypoints~\cite{deng2025general} directly to the policy. Depth and point-cloud inputs are common, but thin fabric remains difficult for depth sensing, and geometry-only observations discard texture cues. Recent benchmarks also show persistent sim-to-real gaps for cloth perception~\cite{hu2025real, ru2025can}.

\textbf{Forward Model.}
A simulator-in-the-loop controller requires a forward model that predicts how cloth responds to candidate actions with sufficient physical fidelity and rollout speed. One direction is to use physical simulators directly. GPU-parallel engines such as Isaac Sim~\cite{NVIDIA_Isaac_Sim}, SAPIEN~\cite{SAPIEN2020Xiang}, Genesis~\cite{Genesis}, and FLASH~\cite{siyuan2026FALSH} provide task-agnostic dynamics and support large-scale rollout, synthetic data generation, and contact-rich interaction. Another direction is to learn dynamics from simulated or real interaction data. For deformable objects, graph-based models such as MeshGraphNets~\cite{pfaff2021meshgraphnets}, DPI-Net~\cite{li2019dpi}, VCD-Cloth~\cite{lin2022vcd}, AdaptiGraph~\cite{zhang2024adaptigraph}, and GraphGarment~\cite{chen2025graphgarment} model mesh, particle, or graph-structured dynamics, while UniClothDiff~\cite{pmlr-v305-tian25c} learns diffusion-based cloth dynamics from synthetic rollouts. These learned models reduce per-rollout cost but remain tied to their training distribution and source simulator.

\textbf{Rollout-based Action Optimization.}
Sampling-based optimization is well suited to contact-rich manipulation because it does not require differentiating through non-smooth dynamics~\cite{frazier2018tutorial, theodorou2010generalized, suh2022differentiable}. CEM~\cite{rubinstein2004cross} and MPPI~\cite{williams2016aggressive, williams2017model} optimize action sequences by evaluating sampled rollouts, and recent variants improve sampling through learned values, covariance adaptation, biased proposals, or diffusion-style annealing~\cite{hansen2022tdmpc, hansen2024tdmpc2, yin2022trajectory, yi2024covompc, trevisan2024biased, xue2025full}. However, most applications focus on rigid-body systems with relatively compact state spaces. In cloth manipulation, high-dimensional deformation and nonconvex geometric objectives make unguided sampling inefficient, often limiting prior work to open-loop primitives or heavily randomized policies~\cite{ha2022flingbot, pmlr-v87-matas18a, jangir2020dynamic}. Closest to our setting, AdaFold~\cite{longhini2024adafold} re-plans folding trajectories in a feedback loop with model predictive control over an RGB-D particle representation, and Kotsovolis and Demiris~\cite{kotsovolis2025garment} drive a predictive controller with a learned garment model. Our controller instead queries the deformable simulator directly, on the mesh state the estimator predicts from a single RGB image.

Other cloth-folding systems obtain corrective behavior through training-time mechanisms rather than inference-time simulator rollout~\cite{longhini2025unfolding}. FoldNet~\cite{chen2026foldnet} synthesizes assets and recovery demonstrations, SPARS~\cite{he2026spars} shapes a progress-aware reward, BiFold~\cite{barbany2025bifold} learns end to end under language guidance, and Hietala et al.~\cite{hietala2022learning} train a visual-feedback policy in simulation with randomized materials. In contrast, our framework uses the full deformable simulator itself as the inference-time forward model after RGB state synchronization.
\section{Method}
\label{sec:method}
In this section, we detail the implementation of the proposed framework illustrated in Fig.~\ref{fig:pipeline}. We begin with the closed-loop execution formulation and notation, then describe the simulator backbone, real-to-sim reconstruction module, and prior-guided MPPI controller.

\subsection{System Overview}
\label{sec:method_overview}
At each control step $t$, the robot receives a single RGB observation $I_t$ of the scene. The real-to-sim module reconstructs a simulation-compatible cloth state
\begin{equation}
    \hat{s}_t = f_{\theta}(I_t),
    \label{eq:real_to_sim}
\end{equation}
which is used to synchronize the simulator. Given a nominal action sequence $\bar{U}_t$ from an offline policy, the controller samples candidate sequences, evaluates them through parallel rollouts in \textsc{FLASH}, and returns a refined action sequence
\begin{equation}
    U_t^{\star} = \mathrm{MPPI}\left(\hat{s}_t, \bar{U}_t, \mathcal{F}, c \right),
    \label{eq:mppi_overview}
\end{equation}
where $\mathcal{F}$ is the simulator rollout operator and $c$ the task cost. The robot executes only its first element, $a_t^{\star}=U_t^{\star}[0]$. The executed action yields a new observation $I_{t+1}$, and the process repeats as a receding-horizon real-to-sim-to-real loop.

\subsection{Simulator Backbone}
\label{sec:flash_backbone}

In this loop, the simulator plays a dual role: it provides synthetic supervision and defines the simulation-compatible state space for real-to-sim training, and it serves as the inference-time rollout engine for MPPI. Its physical fidelity, numerical stability, and rollout efficiency therefore directly determine the performance of the entire real-to-sim-to-real pipeline.

We compare several deformable simulators in Section~\ref{sec:exp_simulator} and build the framework on \textsc{FLASH}~\cite{siyuan2026FALSH}. FLASH represents cloth as a triangular mesh, where vertices are the simulated degrees of freedom and mesh connectivity encodes local deformation and surface geometry. For contact-rich cloth dynamics, FLASH resolves frictional contact through non-smooth Newton iterations on Signorini--Coulomb conditions, preserving stick--slip behavior under large deformation. Its local-global integrator and contact formulation reduce the dominant computation to GPU-friendly sparse matrix operations, enabling stable repeated rollouts for online control~\cite{zeng2025fba}.

Let $x_t=(q_t,\dot{q}_t)$ denote the simulator state, where $q_t$ and $\dot{q}_t$ are the mesh vertex positions and velocities. Given a robot action $a_t$ applied through the end-effector or selected grasping points, FLASH advances the dynamics as
\begin{equation}
    x_{t+1} = \mathcal{F}(x_t, a_t),
    \label{eq:flash_step}
\end{equation}
where $\mathcal{F}$ denotes one simulation step under manipulation, gravity, contact, and friction. During online control, this update is rolled out in parallel over candidate action sequences and passed to MPPI for action refinement.

The estimator predicts positions only, so at each synchronization we set $q_t\!\leftarrow\!\hat{s}_t$ and reset $\dot{q}_t\!\leftarrow\!\mathbf{0}$. Beyond being the only choice consistent with a single RGB frame, this gives all $K$ candidate rollouts an identical initial state, so cost differences reflect the sampled actions alone. It is sound in the quasi-static regime evaluated here.

\subsection{RGB-native Real-to-Sim Reconstruction}
\label{sec:real_to_sim}
We develop the state estimator $f_\theta$ in Equation~\eqref{eq:real_to_sim} to recover a simulator-compatible cloth state $\hat{s}_t$ from a single RGB image. The estimator predicts the deformed positions of $N$ material vertices on the garment asset at each time step $t$:
\begin{equation}
    \hat{s}_t =
    \left\{
    \hat{\mathbf{p}}_{t,i}
    \right\}_{i=1}^{N},
    \qquad
    \hat{\mathbf{p}}_{t,i} \in \mathbb{R}^{3},
    \label{eq:sparse_cloth_state}
\end{equation}
where each vertex corresponds to a fixed material point on the rest-shape mesh constructed offline.
\paragraph{Simulator asset construction.}
For each garment, we reconstruct a fixed-topology rest-shape mesh offline from ZED~2i RGB-D frames. Given a target mesh granularity, we segment the garment with SAM~2~\cite{ravi2025sam}, back-project the masked depth region into a point cloud, triangulate it via ball pivoting~\cite{bernardini1999ball}, and regularize the surface through isotropic remeshing with boundary-locked smoothing~\cite{botsch2004remeshing, taubin1995curve}. This yields a triangular mesh whose material vertices are shared by real-to-sim reconstruction and rollout. Template-based methods such as TRTM~\cite{wang2024trtm} deform a \emph{supplied} template from depth and so presuppose the asset this step builds rather than replacing it.
\paragraph{Synthetic supervision.}
Given the reconstructed asset, we train $f_\theta$ purely on synthetic data generated in \textsc{FLASH}. We simulate manipulation episodes with randomized cloth poses, configurations, and deformation patterns. Every frame is rendered from camera poses matched to the calibrated real setup and paired with the ground-truth vertex positions, giving $M$ image–state pairs:
\begin{equation}
    \mathcal{D}_{\mathrm{syn}}
    =
    \left\{
    \left(
    I^{(m)},
    \left\{
    \mathbf{p}_{i}^{\star(m)}
    \right\}_{i=1}^{N}
    \right)
    \right\}_{m=1}^{M}.
\label{eq:synthetic_realtosim_dataset}
\end{equation}

\paragraph{Network architecture.}
Given the RGB observation $I_t$, a frozen DINOv2 encoder~\cite{oquab2024dinov2} extracts a set of dense patch features, which a learnable linear layer $P$ projects to the model dimension $d$:
\begin{equation}
    V_t = P\!\left(\Phi_{\mathrm{DINO}}(I_t)\right),
    \qquad
    V_t \in \mathbb{R}^{M_v \times d},
    \label{eq:dino_visual_tokens}
\end{equation}
where $\Phi_{\mathrm{DINO}}$ denotes the frozen encoder and $M_v$ is the number of patch tokens. In parallel, we maintain $N$ learnable canonical tokens $C=\{c_i\}_{i=1}^{N}\in\mathbb{R}^{N\times d}$, one per mesh vertex, optimized during training to encode the persistent identity and canonical location of each material point.

The $N$ learnable canonical tokens and $M_v$ visual tokens are concatenated and processed jointly by self-attention layers. Each canonical token aggregates visual evidence from the image while exchanging structural context with others. We take the output slice corresponding to the $N$ canonical tokens:
\begin{equation}
    Z_t =
    \mathrm{SelfAttn}
    \left(
    [\,C\ ;\,V_t\,]
    \right)_{1:N},
    \qquad
    Z_t \in \mathbb{R}^{N \times d},
    \label{eq:query_attention}
\end{equation}
where $[\,\cdot\,;\,\cdot\,]$ denotes token concatenation and $(\cdot)_{1:N}$ extracts the canonical-token outputs. A shared decoder $D_\theta$ maps each token feature to its deformed 3D position:
\begin{equation}
    \hat{\mathbf{p}}_{t,i}
    =
    D_\theta(Z_{t,i}),
    \qquad i=1,\ldots,N.
    \label{eq:point_decoder}
\end{equation}

\paragraph{Training objective.}
We supervise $f_\theta$ against the simulator ground truth with a smooth-$L_1$ loss:
\begin{equation}
    \mathcal{L}_{\mathrm{r2s}}
    =
    \frac{1}{N}
    \sum_{i=1}^{N}
    \ell_{\mathrm{smooth}\text{-}L_1}
    \left(
    \hat{\mathbf{p}}_{i},
    \mathbf{p}^{\star}_{i}
    \right).
    \label{eq:realtosim_loss}
\end{equation}
We optimize only the projection layer $P$, the learnable canonical tokens $C$, the attention layers, and the decoder $D_\theta$, while keeping DINOv2 frozen.
\paragraph{Robust augmentation.}
To improve robustness to sim-to-real visual shift, we inject perturbations at two levels during training. At the input level, we apply image-space randomization to the rendered RGB. At the latent level, we add noise and masking to the visual tokens, including random grasping-point masking around sampled contact regions, so the model learns to infer the cloth state from incomplete and corrupted visual evidence.
\textit{Details of dataset configuration, network architecture, randomization ranges, and augmentation parameters are provided in Appendix~\ref{app:real_to_sim_details}, with the per-task episode-reset ranges also listed in Appendix~\ref{app:exp_state_estimation}}.

\subsection{Prior-Guided MPPI Refinement}
\label{sec:mppi}
We assume access to a base policy that maps the current cloth state to robot actions. At deployment, the base policy produces a nominal action sequence $\bar{U}_t$ over horizon $H$. Rather than being executed directly, this sequence serves as a task-level prior and is refined online through simulator rollouts.

Concretely it is a $512/256/128$ MLP that maps the sparse cloth state to end-effector actions, distilled per garment inside \textsc{FLASH} by on-policy imitation of the scripted teacher, and $\bar{U}_t$ is that policy unrolled for $H$ steps from $\hat{s}_t$ in the simulator.

\paragraph{Prior-guided online refinement.}
Given the synchronized cloth state $\hat{s}_t$, we sample $K$ candidate action sequences around the nominal proposal:
\begin{equation}
    U_t^{(k)}
    =
    \bar{U}_t + \epsilon^{(k)},
    \qquad
    \epsilon^{(k)} \sim \mathcal{N}(0,\Sigma),
    \qquad
    k=1,\ldots,K,
    \label{eq:mppi_sampling}
\end{equation}
where $\Sigma$ controls the local exploration scale. Each candidate is evaluated by rolling it out in the simulator backend:
\begin{equation}
    \tau_t^{(k)}
    =
    \mathcal{F}_{0:H}
    \left(
    \hat{s}_t,
    U_t^{(k)}
    \right),
    \label{eq:mppi_rollout}
\end{equation}
and assigned a task cost $S(\tau_t^{(k)})$ based on the folding objective and regularization terms such as workspace limits, table penetration, and action smoothness.

MPPI converts the rollout costs into normalized weights:
\begin{equation}
    w_k =
    \frac{
    \exp(-S(\tau_t^{(k)})/\lambda)
    }{
    \sum_{j=1}^{K}
    \exp(-S(\tau_t^{(j)})/\lambda)
    },
    \label{eq:mppi_weight}
\end{equation}
where $\lambda$ is the temperature. The refined action sequence is computed as
\begin{equation}
    U_t^{\star}
    =
    \sum_{k=1}^{K}
    w_k U_t^{(k)}.
    \label{eq:mppi_update}
\end{equation}

\paragraph{Receding-horizon execution.}
The optimized sequence is applied in a receding-horizon manner: the robot executes $a_t^{\star}=U_t^{\star}[0]$, receives a new RGB observation, and repeats the real-to-sim-to-real loop. In this way, the base policy is continuously corrected by online physics inference, allowing the system to recover from local execution errors and cloth-state deviations.


\section{Experiments}
\label{sec:experiments}
We organize the evaluation around the design questions in Section~\ref{sec:intro}, validating the necessity of each component and demonstrating the effectiveness of the developed simulator-in-the-loop framework on real-world cloth manipulation tasks.
\subsection{Simulator Backbone}
\label{sec:exp_simulator}
\paragraph{Baselines.}
\setlength{\columnsep}{8pt}
\setlength{\intextsep}{1pt}
\begin{wraptable}{r}{0.7\columnwidth}
\centering
\footnotesize
\setlength{\tabcolsep}{3pt}
\setlength{\abovecaptionskip}{2pt}
\caption{\textbf{Simulator backbone evaluation} under the same cloth manipulation task and MPPI controller.}
\label{tab:simulator_eval}
\begin{tabular}{lcccccc}
\toprule
\multirow{2}{*}{Simulator}
& \multirow{2}{*}{$K$}
& \multicolumn{2}{c}{Time $\downarrow$}
& KP err. $\downarrow$
& MSE $\downarrow$
& \multirow{2}{*}{SR $\uparrow$} \\
\cmidrule(lr){3-4}
& & Step & MPPI  & (cm) & (cm$^2$) & \\
\midrule
\multirow{4}{*}{\textsc{Newton}}
& 64  & 41.0  & 0.85 & $8.40 \pm 8.72$   & $56.8 \pm 73.1$  & 40\% \\
& 128 & 55.1  & 1.14 & $10.28 \pm 11.08$ & $68.9 \pm 93.7$  & 40\% \\
& 256 & 104.3 & 2.15 & $8.85 \pm 10.71$  & $58.1 \pm 89.6$  & 60\% \\
& 512 & 232.8 & 4.80 & $13.02 \pm 13.06$ & $92.3 \pm 105.3$ & 35\% \\
\midrule
\multirow{4}{*}{\textsc{Isaac Sim}}
& 64  & \textbf{11.7} & \textbf{0.26} & $9.78 \pm 10.74$  & $60.5 \pm 104.4$ & 55\% \\
& 128 & \textbf{22.2} & \textbf{0.50} & $10.88 \pm 11.13$ & $67.4 \pm 108.0$ & 45\% \\
& 256 & \textbf{41.1} & \textbf{0.92} & $7.09 \pm 5.60$   & $32.4 \pm 47.6$  & 55\% \\
& 512 & \textbf{80.7} & \textbf{1.79} & $5.84 \pm 4.73$   & $24.6 \pm 32.8$  & 60\% \\
\midrule
\multirow{4}{*}{\textsc{FLASH}}
& 64  & 19.4  & 0.40 & $\mathbf{1.61 \pm 1.18}$ & $\mathbf{3.7 \pm 4.3}$  & \textbf{100\%} \\
& 128 & 36.0  & 0.75 & $\mathbf{1.54 \pm 0.88}$ & $\mathbf{3.1 \pm 1.9}$  & \textbf{100\%} \\
& 256 & 68.9  & 1.42 & $\mathbf{1.95 \pm 1.82}$ & $\mathbf{5.5 \pm 11.0}$ & \textbf{95\%}  \\
& 512 & 137.3 & 2.84 & $\mathbf{2.08 \pm 1.93}$ & $\mathbf{3.9 \pm 3.2}$  & \textbf{95\%}  \\
\bottomrule
\end{tabular}
\vspace{-1.2em}
\end{wraptable}

We compare representative deformable-object simulators, including \textsc{Genesis}~\cite{Genesis}, \textsc{Newton}~\cite{Newton2025}, \textsc{Isaac Sim}~\cite{NVIDIA_Isaac_Sim}, and \textsc{FLASH}~\cite{siyuan2026FALSH}, using the same cloth manipulation task as a controlled probe. The cloth asset, task cost, rollout horizon, MPPI controller, and compute environment are identical.
\paragraph{Metrics.}
We evaluate each simulator in the same MPPI loop under different numbers of parallel environments. For rollout efficiency, we report the wall-clock time per deformable dynamics step (ms) and per MPPI rollout batch (s). For control quality, we report the task-keypoint alignment error (KP err), which is the planner's geometric objective, and define success by whether this error falls below a task-specific threshold. To assess physically implausible final states beyond keypoint alignment, we also report mean per-vertex MSE, Chamfer Distance (CD), and Earth Mover's Distance (EMD) against the target cloth state.

\paragraph{Results.}
Quantitative results are summarized in Table~\ref{tab:simulator_eval} and Fig.~\ref{fig:k_quality_boxplots}. Across all parallel-environment counts, FLASH achieves the best overall trade-off among control accuracy, computation time, and final-state fidelity. Isaac Sim is marginally faster per step (1.5--1.9$\times$), but FLASH attains 95--100\% success across all $K$ while Isaac Sim and Newton stay below 60\%, with per-vertex MSE and EMD roughly an order of magnitude lower, so Isaac Sim falls short on rollout fidelity. Newton further shows occasional numerical instabilities, reflected in its large standard deviations, and Genesis is omitted as it fails to produce a stable fold across all $K$ and seeds, losing table friction under drag. The failure modes are therefore distinct. \textit{Details are provided in Appendix~\ref{app:exp_simulator}.}

We replay \emph{real} end-effector trajectories through \textsc{FLASH} open-loop from a single cloth initialization, scoring the mesh against the recorded cloth with the protocol of~\cite{blancomulero2024benchmarking}. Over $5$ trajectories and $626$ frames the median error is $13.6$~mm against a $5.2$~mm measurement floor, so it is an end-to-end replay-plus-registration discrepancy rather than simulation error alone. Details are provided in Appendix~\ref{app:reb_fidelity}.

\subsection{State Estimation}
\label{sec:exp_state}
\begin{figure}[t]
    \centering
    \includegraphics[width=\linewidth]{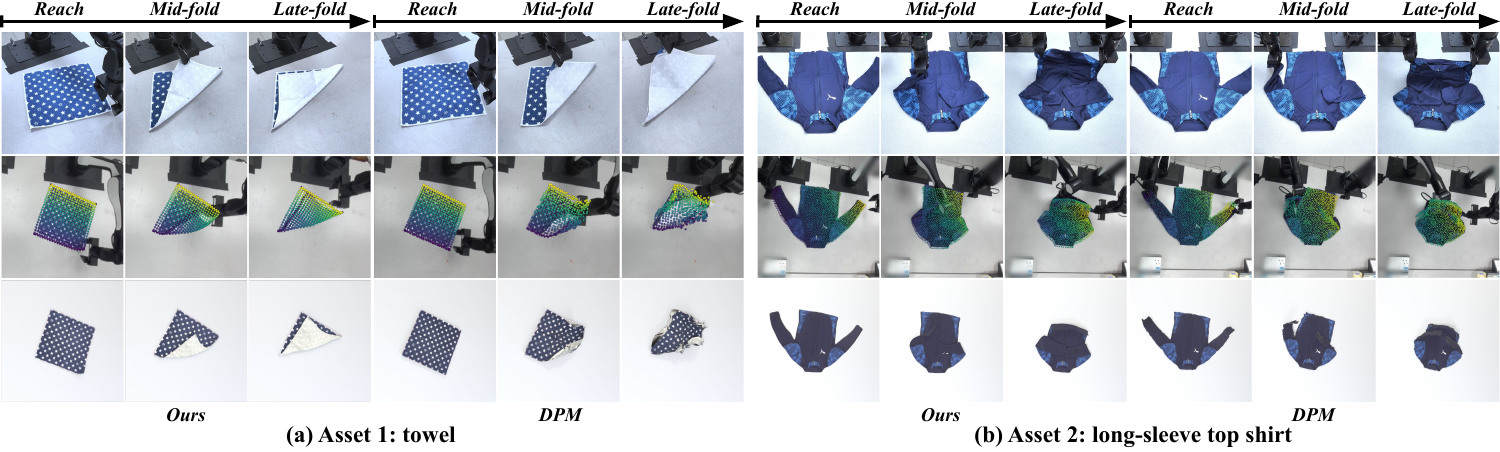}
    \caption{\textbf{Real-world real-to-sim reconstruction.} Panel (a) towel and (b) long-sleeve top, each comparing our RGB-native estimation with DPM at different folding stages. From top to bottom, the real observation, predicted vertices overlaid on the image, and the reconstructed cloth.}
    \label{fig:r2s_overlay}
\vspace{-.5cm}
\end{figure}

\paragraph{Baselines.}
We evaluate on two cloth assets, a towel and a long-sleeve shirt. Baselines include a point-cloud DPM state estimator from UniClothDiff~\cite{pmlr-v305-tian25c} with $50$ denoising steps, and a depth-based variant using DeFM~\cite{patel2026defm} in place of our RGB encoder.
\paragraph{Metrics.}
We report mean per-vertex MSE, CD, and EMD against the simulator ground truth, following~\cite{pmlr-v305-tian25c}. Single-frame inference latency is also recorded for inference efficiency.
\paragraph{Results.}
Table~\ref{tab:state_estimation} summarizes the quantitative comparison. Our RGB-native module achieves the lowest reconstruction error on both assets across all three metrics while maintaining the fastest inference, showing stronger material-point correspondence and better preservation of the overall cloth shape. The per-vertex MSE is 24.23~mm$^2$ on the long-sleeve shirt and 2.33~mm$^2$ on the towel (about 4.9 vs.\ 1.5~mm in RMS), while the CD and EMD remain within a few millimeters and the lowest among all methods. Fig.~\ref{fig:r2s_overlay} further visualizes real-world deployment examples. Poor vertex accuracy or weak shape preservation can produce abnormal real-to-sim assets, whereas our method maintains both material-point consistency and global cloth shape, enabling faithful simulator initialization for rollout.

The absolute DPM values are not comparable to those in~\cite{pmlr-v305-tian25c}, since we retrain it on our garments, topology and fold distribution. Scored against \emph{real} depth back-projections rather than simulator ground truth, our estimator reaches $10.2\pm1.4$~mm per trajectory.
\textit{Baseline details and both analyses are provided in Appendix~\ref{app:exp_state_estimation},~\ref{app:reb_details}}.

\begin{table}[t]
    \centering
    \footnotesize
    \setlength{\tabcolsep}{6pt}
    \caption{\textbf{State-estimation comparison.} Per-vertex MSE, CD, EMD, and single-frame inference latency on the validation split.}
    \label{tab:state_estimation}
    \begin{tabular}{llcccc}
        \toprule
        Asset & Real-to-Sim & MSE (mm$^2$) $\downarrow$ & CD (mm) $\downarrow$ & EMD (mm) $\downarrow$ & Lat. (ms) $\downarrow$ \\
        \midrule
        \multirow{3}{*}{Towel}
            & \textbf{Ours} & \cellcolor{ourblue}$\mathbf{2.33 \pm 1.21}$  & \cellcolor{ourblue}$\mathbf{1.94 \pm 0.41}$ & \cellcolor{ourblue}$\mathbf{1.98 \pm 0.45}$ & \cellcolor{ourblue}$\mathbf{7.4 \pm 0.6}$ \\
            & DPM   & $7.17 \pm 2.22$           & $3.89 \pm 0.67$          & $4.01 \pm 0.72$          & $318.5 \pm 3.7$ \\
            & DeFM        & $43.36 \pm 34.66$         & $5.51 \pm 0.99$          & $6.53 \pm 1.52$          & $12.6 \pm 0.5$ \\
        \midrule
        \multirow{3}{*}{\makecell{Long-sleeve\\Shirt}}
            & \textbf{Ours} & \cellcolor{ourblue}$\mathbf{24.23 \pm 11.49}$ & \cellcolor{ourblue}$\mathbf{3.65 \pm 0.78}$ & \cellcolor{ourblue}$\mathbf{4.67 \pm 1.24}$ & \cellcolor{ourblue}$\mathbf{12.4 \pm 0.4}$ \\
            & DPM  & $45.74 \pm 14.19$          & $5.93 \pm 0.27$          & $8.21 \pm 0.65$          & $512.9 \pm 6.0$ \\
            & DeFM        & $43.52 \pm 22.02$          & $4.05 \pm 0.38$          & $5.93 \pm 0.91$          & $18.0 \pm 0.6$ \\
        \bottomrule
    \end{tabular}
\end{table}

\subsection{Real-World Evaluation}
\label{sec:exp_realworld}

\paragraph{Tasks.}
We construct end-to-end pipelines with different module choices and evaluate them on two real-world tasks: \textit{single-arm diagonal folding} and \textit{dual-arm symmetric folding}. For each trial, we randomize the initial towel pose and run the method on the same robot platform.
\paragraph{Baselines.}
We evaluate variants by replacing the state estimator, simulator backbone, or controller in the full pipeline. Specifically, we compare the RGB-native module with DPM~\cite{pmlr-v305-tian25c}, the \textsc{FLASH} backend with Isaac Sim~\cite{NVIDIA_Isaac_Sim}, and prior-guided MPPI refinement with direct base-policy execution.
A trial is successful if every target corner pair is aligned within $2$~cm in the final cloth state. A trial succeeds only if the grasp succeeds \emph{and} the alignment criterion is met, so all reported rates already include grasping and no trial was discarded or re-run. Grasp success is reported separately in Appendix~\ref{app:exp_realworld}.

\paragraph{Results.}
\begin{wraptable}{r}{0.6\linewidth}
    \centering
    \footnotesize
    \setlength{\tabcolsep}{4pt}
    \setlength{\abovecaptionskip}{2pt}
    \caption{\textbf{Real-world pipeline-variant evaluation.} Each row replaces one component of the full pipeline (first row).}
    \label{tab:realworld}
    \begin{tabular}{llllc}
        \toprule
        Task & Real-to-Sim & Backend & Controller & Real SR ($\uparrow$) \\
        \midrule
        \multirow{4}{*}{\makecell[l]{Single-arm \\ diagonal}}
            & \cellcolor{ourblue}\textbf{Ours (RGB)} & \cellcolor{ourblue}\textbf{FLASH}     & \cellcolor{ourblue}\textbf{MPPI}      & \cellcolor{ourblue}\textbf{9/10} \\
            & DPM                 & FLASH              & MPPI               & 6/10          \\
            & Ours (RGB)          & Isaac Sim          & MPPI               & 5/10          \\
            & Ours (RGB)          & FLASH              & Base-policy        & 3/10          \\
        \midrule
        \multirow{4}{*}{\makecell[l]{Dual-arm \\ symmetric}}
            & \cellcolor{ourblue}\textbf{Ours (RGB)} & \cellcolor{ourblue}\textbf{FLASH}     & \cellcolor{ourblue}\textbf{MPPI}      & \cellcolor{ourblue}\textbf{8/10} \\
            & DPM                 & FLASH              & MPPI               & 3/10          \\
            & Ours (RGB)          & Isaac Sim          & MPPI               & 4/10          \\
            & Ours (RGB)          & FLASH              & Base-policy        & 1/10          \\
        \bottomrule
    \end{tabular}
\end{wraptable}
The full pipeline achieves the highest success rate on both tasks, demonstrating a substantial closed-loop improvement over direct base-policy execution. The drop in performance from replacing any single component further shows that state synchronization, simulator fidelity, and online refinement are tightly coupled, and that weakening any part degrades the entire real-to-sim-to-real loop.
\textit{All variants use the same robot platform, camera setup, action primitive, task cost, and stopping criterion. Detailed configurations and additional trial visualizations are provided in Appendix~\ref{app:exp_realworld}.}

\subsection{Generalization and Closed-Loop Correction}
\label{sec:exp_realworld_generalization}
\paragraph{Tasks.}
\begin{wraptable}{r}{0.47\linewidth}
    \centering
    \footnotesize
    \setlength{\tabcolsep}{5pt}
    \setlength{\abovecaptionskip}{2pt}
    \caption{\textbf{Generalization and disturbance-recovery evaluation.} }
    \label{tab:generalization}
    \begin{tabular}{lc}
        \toprule
        Task & Real SR \\
        \midrule
        Long-sleeve     & 8/10  \\
        Shorts                   & 7/10  \\
        Towel, reverse diag.     & 8/10 \\
        Towel, mid-fold disturb. & 8/10 \\
        \bottomrule
    \end{tabular}
\end{wraptable}
We further evaluate the framework on four real-world settings: long-sleeve shirt folding, shorts folding, reverse-diagonal towel folding, and mid-fold disturbance recovery. These settings test generalization across garment assets and folding variants, as well as closed-loop correction under external perturbations. All tasks follow the same evaluation protocol as the towel experiments, with task-specific geometric success criteria. Full task configurations are provided in Appendix~\ref{app:exp_generalization}.

These settings vary geometry and topology ($464$-vertex towel to notched $1156$-vertex shorts and $1379$-vertex long-sleeve top), fold sequence, arm count, and the pick and goal keypoints used to distill the prior. Mass, bending, and friction parameters are tuned separately for each asset.

\begin{figure}[t]
    \centering
    \includegraphics[width=0.85\linewidth]{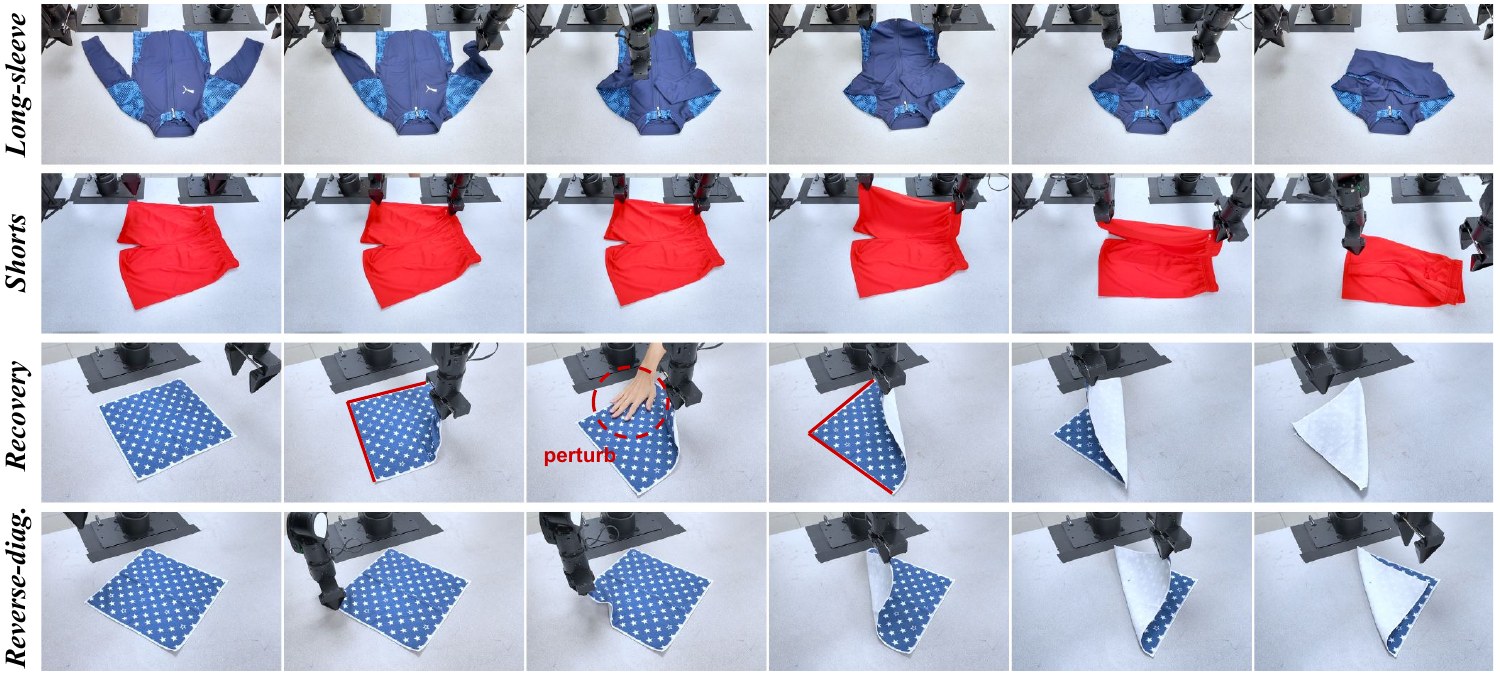}
    \caption{
    \textbf{Representative rollouts for the generalization and disturbance-recovery settings.}
    Time progresses from left to right. The four rows visualize long-sleeve shirt folding, shorts folding, recovery from mid-fold human disturbance, and reverse-diagonal towel folding.
    }
    \label{fig:generalization_results}
\vspace{-.5cm}
\end{figure}

\paragraph{Results.}
\setlength{\columnsep}{8pt}
\setlength{\intextsep}{1pt}
Table~\ref{tab:generalization} reports the success rate over ten trials for each task, and Fig.~\ref{fig:generalization_results} shows one representative execution sequence per setting. The framework remains reliable across new garment assets, folding variants, and mid-fold disturbances without additional dynamics retraining, showing that the same loop transfers across task and asset variations and corrects for perturbations encountered during execution.

Ten binary trials do not show whether the error is actually driven down, so we also measure the correction over $20$ seeds in simulation (Appendix~\ref{app:reb_closedloop}). There, $18/20$ seeds reach the $2$~cm criterion, reducing the keypoint error from $42.2$~cm at grasp to $0.8\pm0.1$~cm in $22$--$31$ steps. Pushing the cloth $6$, $9$, or $12$~cm eight steps into the fold, the post-disturbance error returns below its pre-disturbance level after $4.0\pm0.0$, $4.1\pm0.7$, and $6.0\pm1.0$ steps.

\section{Conclusion}
\label{sec:conclusion}
 We presented a simulator-in-the-loop framework that brings inference-time physics refinement to real-world cloth manipulation from a single RGB observation, coupling the FLASH deformable simulator, an RGB-native real-to-sim module, and a prior-guided MPPI controller into one closed-loop real-to-sim-to-real pipeline. Real-robot experiments show that the framework attains higher success rates than baselines, and closed-loop correction under mid-fold perturbations.


\section{Limitations and Future Work}
\label{sec:limitations}
 First, although our system strikes a good balance between rollout efficiency and physical fidelity, solving deformable dynamics is far less parallelizable than the rigid-body case, so rollouts easily saturate the single-GPU budget, and the achievable batch size, which upper-bounds the closed-loop control rate, remains tied to the simulation backend. Future work will scale inference across multiple GPUs via distributed rollout and further backend optimization. Second, generalization to novel garments is still limited by two per-asset stages: the offline cloth-mesh reconstruction and the real-to-sim module, which is currently trained per garment. The shirt and the shorts each use a separately trained estimator. Adding a garment costs one RGB-D capture, a scripted meshing pass, ${\sim}1$~h of data generation and ${\sim}12$~h of training on one RTX~4090, while the backend, cost, controller and deployment stack are shared, so a garment costs one retrained module rather than a new system. Generalizing that module across garments is the main open problem for this line of work. We see a promising direction in image-conditioned mesh foundation models that produce a simulator-compatible asset from a single observation, paired with a category-level real-to-sim module that removes the per-garment training step.
The estimator observes position but not velocity, so synchronization resets vertex velocities to zero, which bounds the method to quasi-static manipulation. Material class does not vary across the evaluated garments, which are woven textiles of comparable thickness. The scope stays within tabletop folding under a fixed overhead camera.


\clearpage
\acknowledgments{This work was supported by the MOE AcRF Tier 1 24-1234-P0001 Funding, NUS Presidential Young Professorship, and Google Gift Fund. The authors would also like to acknowledge the support of the NUS AI Institute.}


\bibliography{main}  

\clearpage

\renewcommand{\thesection}{\Alph{section}}
\renewcommand{\thefigure}{\Alph{section}.\arabic{figure}}
\renewcommand{\thetable}{\Alph{section}.\arabic{table}}
\counterwithin*{figure}{section}
\counterwithin*{table}{section}
\setcounter{section}{0}
\setcounter{figure}{0}
\setcounter{table}{0}
\renewcommand{\theHsection}{app.\Alph{section}}
\renewcommand{\theHsubsection}{app.\Alph{section}.\arabic{subsection}}

\section*{\Large Appendix}
\section{Implementation Details}
\label{app:implementation}

    This section provides the full implementation details of the two modules introduced in the main paper. We describe the RGB-native real-to-sim reconstruction module in \S\ref{app:real_to_sim_details} and the prior-guided MPPI controller, including its cost terms and hyperparameters, in \S\ref{app:mppi}. \S\ref{app:modeling_assumptions} states the modeling assumptions that both rely on.

\subsection{RGB-native Real-to-Sim Reconstruction}
\label{app:real_to_sim_details}

    The real-to-sim module maps a single RGB frame to the deformed 3D positions of a fixed set of $N$ cloth mesh vertices, the same vertex set the planner rolls out. The model has four trainable components on top of a frozen visual backbone: a linear projection $P$, a bank of $N$ learnable canonical tokens, a self-attention fusion backbone, and a per-vertex MLP decoder.

    \textbf{Input.}
    The input is a single RGB frame center-cropped and resized to $392\!\times\!392$ and normalized with the standard ImageNet statistics. The same crop and normalization are applied identically at training and test time.

    \textbf{Visual Encoder.}
    The encoder is a frozen pretrained DINOv2-small~\cite{oquab2024dinov2} ViT-S/14, with patch size $14$. The $392\!\times\!392$ input yields $M_v=(392/14)^2=784$ dense patch tokens at the backbone hidden dimension $d_{\mathrm{vit}}=384$. A learnable linear projection $P\!:\mathbb{R}^{384}\!\to\!\mathbb{R}^{d}$ maps each patch token to the model dimension $d=256$.

    \textbf{Canonical Tokens.}
    We maintain $N$ learnable canonical tokens $C=\{c_i\}_{i=1}^{N}\in\mathbb{R}^{N\times d}$, one per mesh vertex ($N=464$ for the towel, $N=1156$ for the pants, and $N=1379$ for the long-sleeve top). Each $c_i$ is initialized from a deterministic Gaussian random Fourier-feature map of its rest-shape coordinate $\mathbf{x}_i^{\mathrm{rest}}\in\mathbb{R}^3$:
    \begin{equation}
        c_i^{(0)}
        =
        \big[\,\sin(2\pi\,\mathbf{x}_i^{\mathrm{rest}} B);\,\cos(2\pi\,\mathbf{x}_i^{\mathrm{rest}} B)\,\big]
        \in \mathbb{R}^{d},
        \qquad B \in \mathbb{R}^{3\times d/2},\ B_{jk}\sim\mathcal{N}(0,\sigma_{B}^{2}),
        \label{eq:app_fourier_init}
    \end{equation}
    with $\sigma_B=4.0$ and a fixed random seed. The matrix $B$ is sampled once and frozen, while $C$ is optimized as a free parameter table during training.

    \textbf{Fusion Backbone.}
    The $N$ canonical tokens are concatenated with the $M_v$ projected visual tokens into a sequence of length $N+M_v$ and processed by a stack of $L=4$ standard pre-norm self-attention blocks (multi-head self-attention $+$ feed-forward) operating at $d=256$ with $h=8$ heads and dropout $0.1$. The output slice corresponding to the first $N$ positions is taken as the fused canonical features $Z\in\mathbb{R}^{N\times d}$. Structural context and visual evidence are exchanged jointly by self-attention.

    \textbf{Decoder.}
    A shared per-vertex MLP decoder maps each $Z_i$ to its predicted 3D position. It applies a LayerNorm followed by two hidden Linear$\to$GELU blocks of width $512$ and a final Linear projection to $\mathbb{R}^{3}$:
     \begin{equation}
        \hat{\mathbf{p}}_i = W_{\mathrm{out}}\,\mathrm{GELU}(W_{2}\,\mathrm{GELU}(W_{1}\,\mathrm{LN}(Z_i))),
        \label{eq:app_decoder}
    \end{equation}
    where $W_1\in\mathbb{R}^{256\times 512}$, $W_2\in\mathbb{R}^{512\times 512}$, and $W_{\mathrm{out}}\in\mathbb{R}^{512\times 3}$.

    \textbf{Loss.}
    We supervise the model against the simulator ground-truth vertex positions with a smooth-$L_1$ loss (Huber threshold $\beta=1.0$), reduced as the mean over the $N$ vertices and the batch.

    \textbf{Training Data.}
    Paired RGB--mesh examples are generated in \textsc{FLASH} by replaying scripted goal-guided demonstration trajectories and rendering each frame from the camera pose that matches our real-world setup. For every frame we record the ground-truth deformed 3D positions of all $N$ cloth-mesh vertices. The dataset is partitioned $90\%/10\%$ between training and validation at the episode level so that no frame leaks across the split.

    \textbf{Randomization Ranges.}
    Episode-level randomization of the cloth pose is applied at every reset. The cloth centroid offset is drawn as $\Delta x \sim \mathcal{U}(0.16, 0.40)$~m and $\Delta z \sim \mathcal{U}(-0.40, -0.20)$~m in the table plane, and the yaw angle as $\sim \mathcal{U}(-0.50, 0.50)$~rad. These are the ranges also used for the state-estimation evaluation of \S\ref{app:exp_state_estimation}. Deformation is not controlled by an explicit parameter range. It comes from gravity-drop states, in which the cloth is released and allowed to settle, together with every intermediate frame of the scripted folds. Image-level and token-level augmentation ranges are tabulated separately in Tab.~\ref{tab:app_rgb_aug}.

    \textbf{Augmentation.}
    Two-level perturbations are applied at training time to bridge the sim-to-real visual gap: photometric, geometric, and lighting randomization on the RGB input, plus token-level masking that propagates a grasp-point pixel occlusion into the overlapping DINOv2 patch tokens. The full parameter ranges are tabulated in Tab. \ref{tab:app_rgb_aug}.
        \begin{table}[htbp]
        \centering
        \caption{\textbf{Training-time augmentation for the RGB module.} Every randomized operation applied during training, with its application probability and parameter range. Color jitter and Gaussian blur run on every frame with parameters sampled from the listed ranges. The grasp-point token mask is propagated only when the grasp-point pixel occlusion is applied, so its probability is conditional on that event.}
        \label{tab:app_rgb_aug}
        \small
        \setlength{\tabcolsep}{5pt}
        \begin{tabular}{lllp{0.5\linewidth}}
            \toprule
            Operation & Level & Prob. & Range / strength \\
            \midrule
            Color jitter           & Image & $1.0$ & brightness, contrast, saturation $\pm0.4$, hue $\pm0.1$ \\
            Random grayscale       & Image & $0.2$ & convert RGB to single channel \\
            Gaussian blur          & Image & $1.0$ & kernel $5$, $\sigma \in [0.1, 2.0]$ \\
            Directional lighting   & Image & $0.6$ & gradient depth $\alpha \in [0.15, 0.40]$ \\
            Random erasing         & Image & $0.3$ & erased area fraction $\in [0.02, 0.15]$ \\
            Grasp-point occlusion  & Image & $0.7$ & half-side $\in [25, 55]$ px, aspect $\in [0.7, 1.5]$, corner jitter $\pm12$ px, center jitter $\pm10$ px, fill brightness $\in [0, 0.05]$ \\
            Grasp-point token mask & Token & $0.8$ & DINOv2 patches with $> 30\%$ area overlap flagged invalid for fusion \\
            \bottomrule
        \end{tabular}
    \end{table}

    \textbf{Optimizer and Schedule.}
    The trainable parameters are $\{P,\,C,\,\text{self-attn},\,\text{decoder}\}$; the DINOv2 backbone is frozen. We use AdamW with learning rate $3\!\times\!10^{-4}$, weight decay $10^{-4}$, and gradient-norm clipping at $1.0$. The learning rate follows a cosine schedule with a $3$-epoch linear warmup and a minimum-LR ratio of $10^{-2}$. Training runs for $50$ epochs with a batch size of $32$ on a single NVIDIA RTX~4090.

    \textbf{Deployment.}
    At test time, each RGB frame is center-cropped to $392\!\times\!392$ and passed through the network to obtain the predicted 3D positions of all $N$ vertices. The predictions are transformed from camera frame to world frame and broadcast into all $K$ parallel simulation environments. The end-effector pose and gripper state are read from the robot proprioception stack and synchronized in the same step.

\subsection{Prior-Guided MPPI Control}
\label{app:mppi}

    \textbf{Prior-Informed Sampling.}
    At each planning step, the cloth state is synchronized from the real-to-sim module (\S\ref{app:real_to_sim_details}) into all $K$ parallel simulation environments. Our default MPPI configuration uses $K = 256$ parallel sampled trajectories, a planning horizon of $H = 5$ steps, a per-dimension noise standard deviation of $\sigma = 5\!\times\!10^{-3}$~m, and a softmax temperature of $\lambda = 0.08$. The proposal noise is temporally independent Gaussian, $\boldsymbol{\epsilon}_t^{(k)} \sim \mathcal{N}(\mathbf{0}, \sigma^2 \mathbf{I})$, and the $K$ rollouts are dispatched to $K$ parallel simulation environments that remain resident on the GPU across planning steps. The proposal mean is anchored at the distilled-prior nominal $\bar{U}_t$ as described in the main paper, and refined locally by the importance-weighted update.

    \textbf{Cost Function Details.}
    The stage cost combines a task-driving distance term with a small set of safety and smoothness penalties whose mix changes across the manipulation phases. Let $\mathbf{x}_{\mathrm{ee}} \in \mathbb{R}^3$ denote the end-effector position, $z_{\mathrm{ee}}$ its vertical coordinate, and $z_{\mathrm{table}}$ the table surface height. The distance term is phase-dependent. In the \emph{reach} phase it is the full 3D distance from the end-effector to the pick corner $\mathbf{p}_{\mathrm{pick}}$. In the \emph{fold} phase it is the in-plane distance from the grasped cloth corner $\mathbf{p}_{\mathrm{grasp}}$ to the fold target $\mathbf{p}_{\mathrm{target}}$. The remaining components penalize table penetration, workspace-boundary violation, and action magnitude. The terminal cost $\phi(x_H)$ reuses the phase-dependent distance evaluated at the final rolled-out state. Expressions, weights, and activations are listed in Table~\ref{tab:app_cost}.

    \begin{table}[htbp]
        \centering
        \caption{\textbf{MPPI stage cost components.} $\mathbf{x}_{\mathrm{ee}}^{xy}$ denotes the in-plane end-effector position. $\mathcal{W}_{\mathrm{safe}}^{\mathrm{inner}}$ denotes the interior of the safe workspace after shrinking each axis by a soft margin $\delta_{\mathrm{bnd}}$. $z_{\mathrm{fold}}^{*}$ is the nominal lift height during folding.}
        \label{tab:app_cost}
        \begin{tabular}{llcl}
            \toprule
            Term & Expression & Weight & Activation \\
            \midrule
            $c_{\mathrm{dist}}$ (reach) & $\| \mathbf{x}_{\mathrm{ee}} - \mathbf{p}_{\mathrm{pick}} \|_2$                                                               & $1.0$  & \emph{reach} phase \\
            $c_{\mathrm{dist}}$ (fold)  & $\| \mathbf{p}_{\mathrm{grasp}}^{xy} - \mathbf{p}_{\mathrm{target}}^{xy} \|_2$                                                & $1.0$  & \emph{fold} phase \\
            $c_{\mathrm{table}}$        & $\max\!\left(0,\, z_{\mathrm{table}} + \delta_{\mathrm{tbl}} - z_{\mathrm{ee}}\right)^{2}$                                     & $50.0$ & all phases \\
            $c_{\mathrm{limit}}$        & axis-wise squared hinge on $\mathcal{W}_{\mathrm{safe}}^{\mathrm{inner}}$                                                     & $20.0$ & all phases \\
            $c_{\mathrm{vel}}$          & $\| \mathbf{a}^c_t \|_2^2$                                                                                                    & $10.0$ & all phases \\
            $c_{\mathrm{height}}$       & $(z_{\mathrm{ee}} - z_{\mathrm{fold}}^{*})^{2}$                                                                               & $2.0$ & \emph{fold} phase \\
            \bottomrule
        \end{tabular}
    \end{table}

    At deployment we use a table safety margin $\delta_{\mathrm{tbl}} = 5$~mm, and a workspace soft-boundary margin $\delta_{\mathrm{bnd}} = 2$~cm.

\subsection{Modeling Assumptions}
\label{app:modeling_assumptions}

    \textbf{Material parameters and sensitivity.} Cloth mass, bending, and friction are specified per garment and tuned for the solver. The closed-loop design is intended to limit the accumulation of errors from parameter mismatch, but it does not remove sensitivity to these parameters. MPPI uses the simulator to \emph{rank} candidate sequences over a five-step horizon rather than to predict an accurate long-horizon trajectory, and the state is re-synchronized from observation at every step, so parameter error is not integrated forward. For substantially thicker fabrics, an RGB-D front end substitutes into the same pipeline without changing the backend, cost, or controller.

    \textbf{Contact abstraction.} The gripper is modeled as a rigid box that pins the grasped vertices for the duration of the grasp. This is adequate for the corner grasps evaluated here, but tasks requiring sliding, regrasping, or partial grasps would need a contact model that resolves finger--cloth friction explicitly.


\section{Experiment Details}
\label{app:exp_details}

    This section documents the detailed configurations behind the experiments referenced in the main paper. We describe the simulator backbone comparison in \S\ref{app:exp_simulator}, the state-estimation dataset and baseline configurations in \S\ref{app:exp_state_estimation}, the real-world pipeline-variant evaluation in \S\ref{app:exp_realworld}, and the generalization and closed-loop correction study in \S\ref{app:exp_generalization}.

\subsection{Simulator Backbone Comparison}
\label{app:exp_simulator}

\begin{figure}[t]
      \centering
      \begin{subfigure}[t]{0.48\linewidth}
          \centering
          \includegraphics[width=\linewidth]{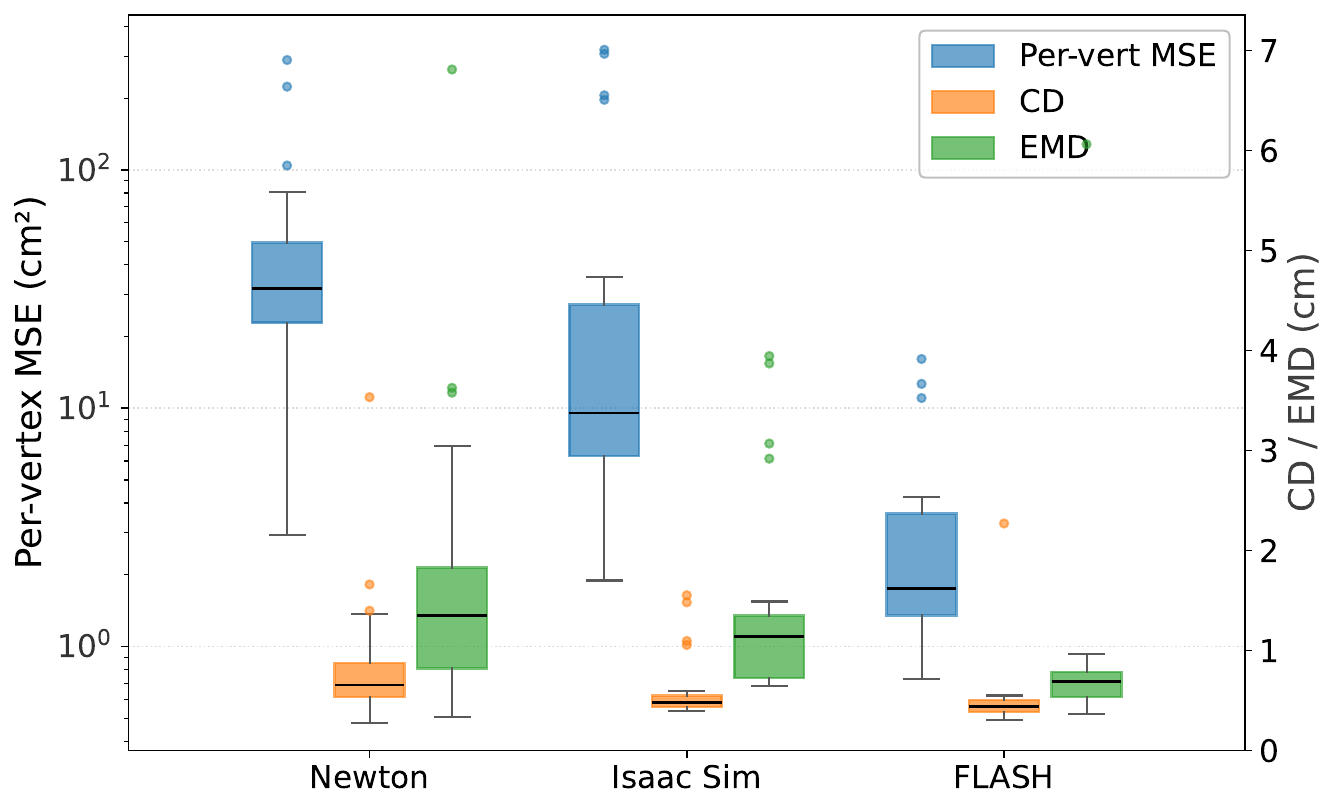}
          \caption{$K=64$}
          \label{fig:k64_box}
      \end{subfigure}
      \hfill
      \begin{subfigure}[t]{0.48\linewidth}
          \centering
          \includegraphics[width=\linewidth]{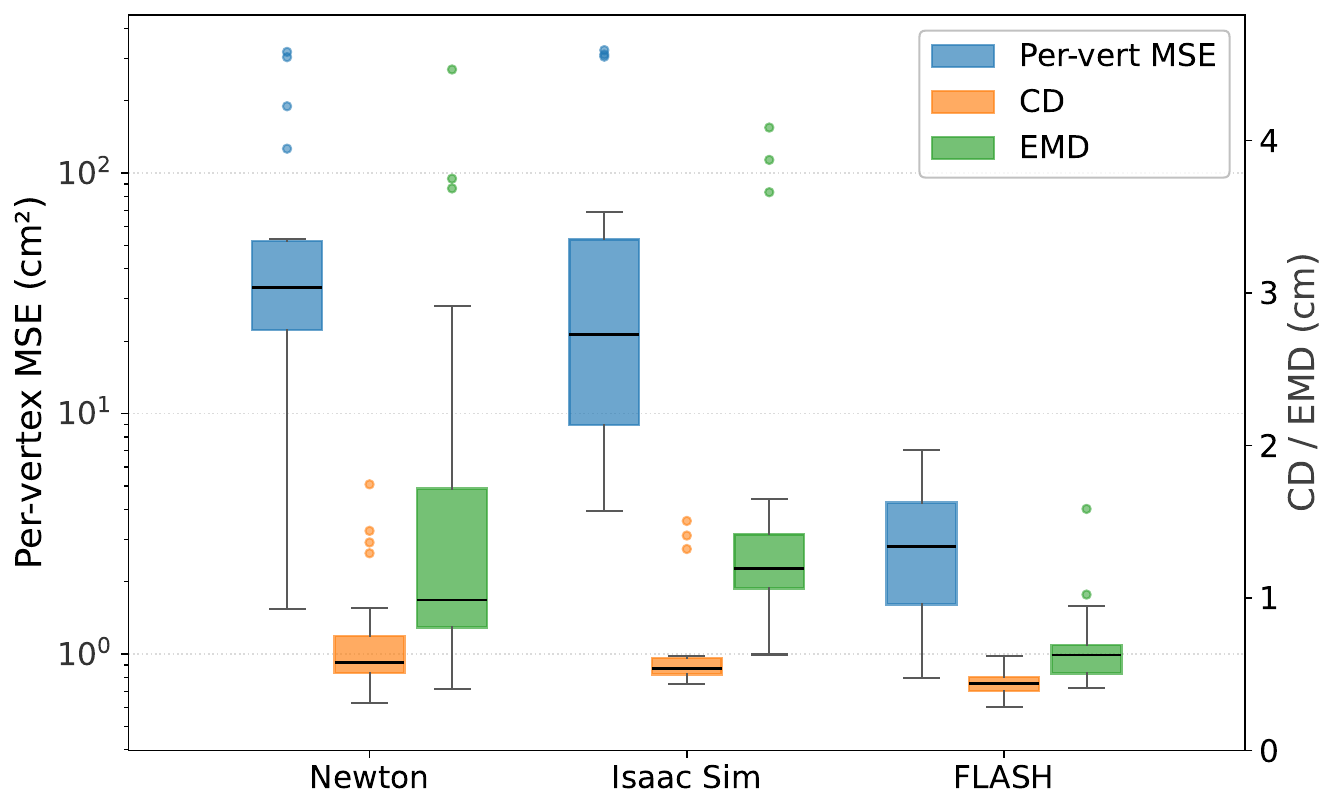}
          \caption{$K=128$}
          \label{fig:k128_box}
      \end{subfigure}
      \\[0.6em]
      \begin{subfigure}[t]{0.48\linewidth}
          \centering
          \includegraphics[width=\linewidth]{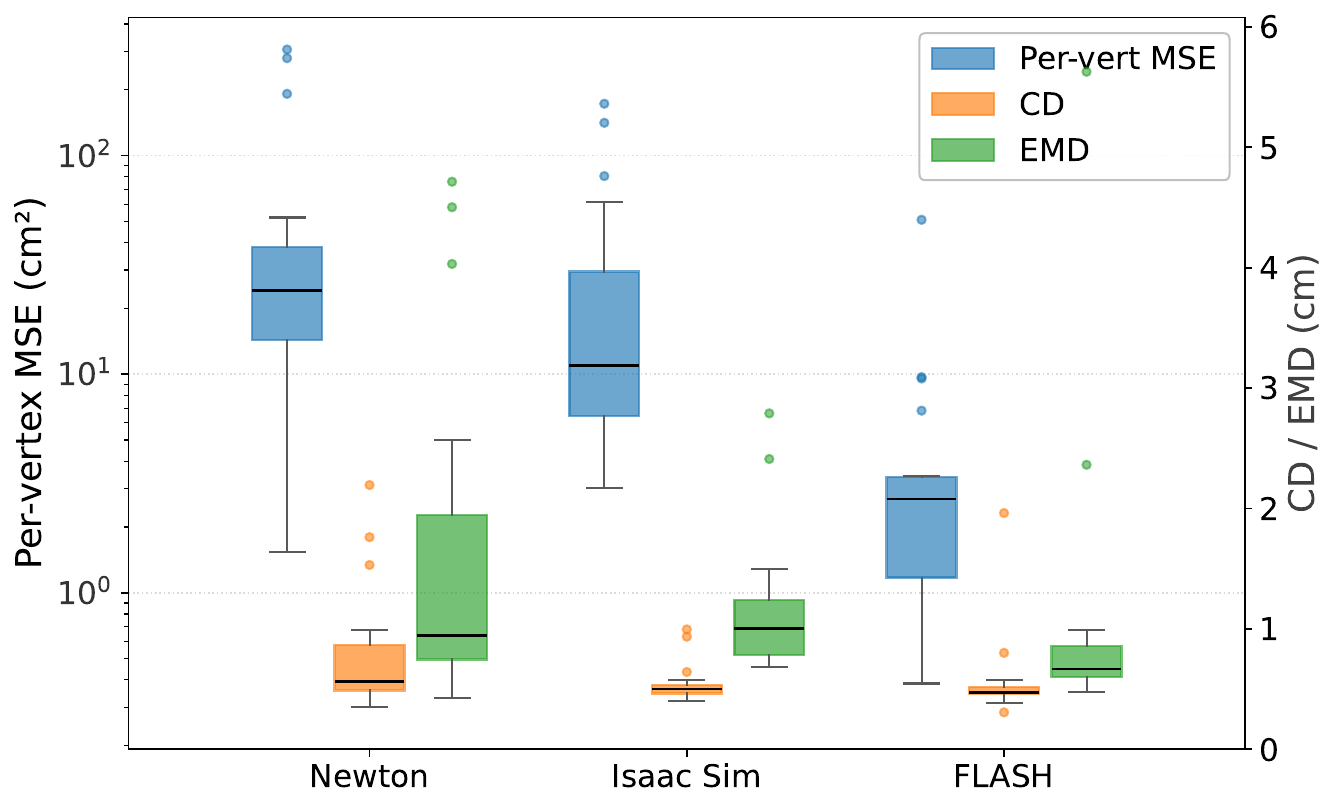}
          \caption{$K=256$}
          \label{fig:k256_box}
      \end{subfigure}
      \hfill
      \begin{subfigure}[t]{0.48\linewidth}
          \centering
          \includegraphics[width=\linewidth]{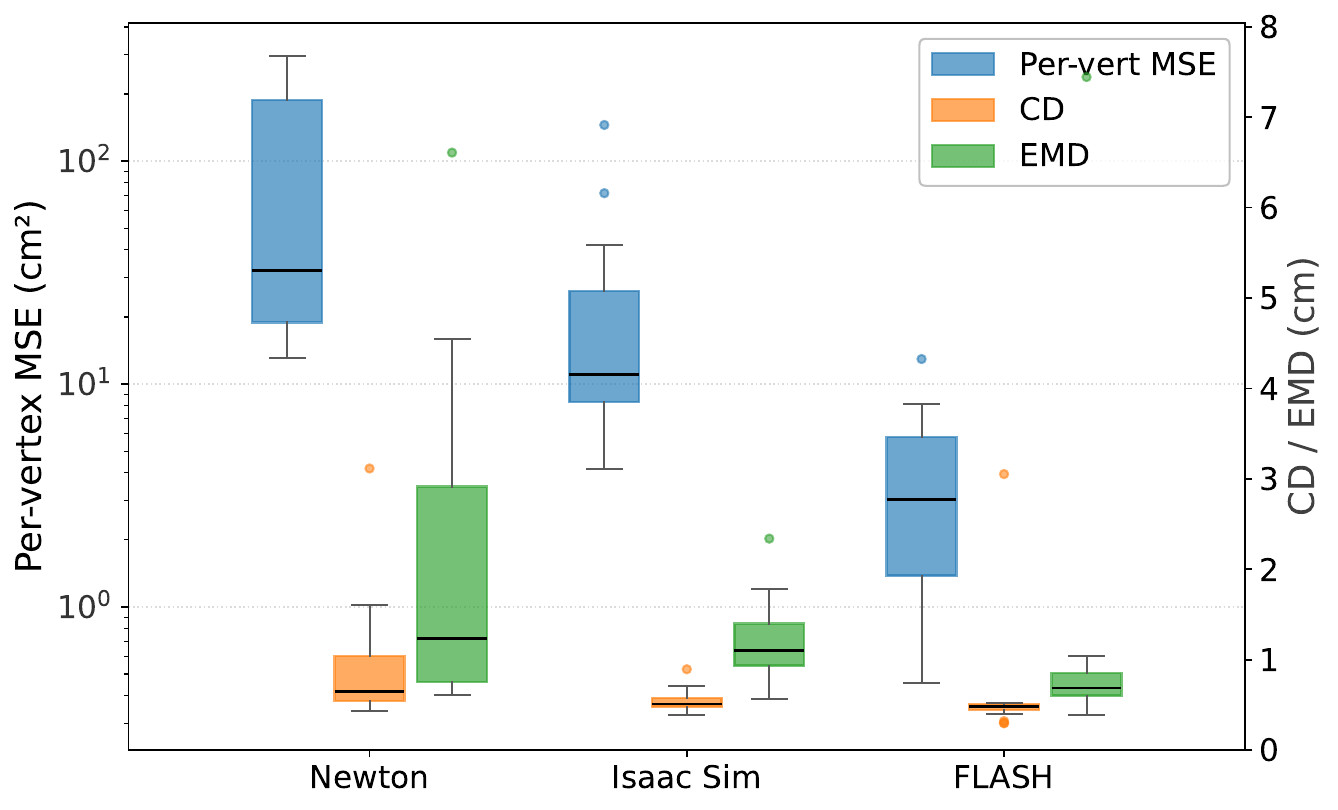}
          \caption{$K=512$}
          \label{fig:k512_box}
      \end{subfigure}
      \caption{\textbf{Settled-state fold-quality distributions across simulators at four parallel environment counts.} Each panel shows the
  20-seed distribution of per-vertex MSE (cm$^2$, log left axis), CD (cm), and EMD (cm) for Newton, Isaac Sim, and FLASH under identical seeded initial cloth poses.}
      \label{fig:k_quality_boxplots}
  \end{figure}

    We compare \textsc{Genesis}~\cite{Genesis}, \textsc{Newton}~\cite{Newton2025}, \textsc{Isaac Sim}~\cite{NVIDIA_Isaac_Sim}, and \textsc{FLASH}~\cite{siyuan2026FALSH} on the same single-arm diagonal-fold task under a shared vanilla MPPI controller, varying only $K \in \{64,128,256,512\}$. Main results are in Table~\ref{tab:simulator_eval} and Fig.~\ref{fig:k_quality_boxplots}.

    \textbf{Task and controller.}
    The cloth is a $30\!\times\!30$~cm towel mesh of $N_{\mathrm{towel}}=464$ vertices. MPPI plans an $H{=}6$-step horizon at $\Delta t_{\mathrm{act}}{=}40$~ms with per-step EE translation clamped to $20$~mm; each simulator runs at $\Delta t_{\mathrm{sim}}{=}10$~ms (i.e., $4$ simulation steps per action). Trials early-stop when KP err.\ $<5$~cm, then settle $150$ further steps to acquire the final folding behavior.

    \textbf{Seeds and compute.}
    Each $K$ uses $20$ fixed seeds over a $\pm 8$~cm $/$ $\pm 30^{\circ}$ init randomization. The same per-seed initial pose is replayed on every simulator to ensure a fair comparison. All trials run on a single RTX~4090.

    \textbf{Metrics.}
    \emph{Step} and \emph{MPPI} are wall-clock per simulation step and per MPPI iteration, respectively. \emph{KP err.}\ is the settled horizontal distance between pick- and target-corner. \emph{MSE} is the mean squared horizontal distance between each moved-half vertex ($231$ on the $464$-vertex mesh) and its mirror partner in the stationary half (cm$^{2}$), which measures cloth overlap after the fold, independent of the rest target. \emph{CD} and \emph{EMD} (Fig.~\ref{fig:k_quality_boxplots}) compare the settled cloth to the ideal perfect-fold triangle. \emph{SR} is the seed fraction with KP err.\ $<5$~cm.

    \textbf{Simulator configurations.}
   Table~\ref{tab:app_exp_simulator_config} lists the cross-simulator settings. Cloth material parameters are omitted from the table because they are not commensurable across solver types. Each cloth was properly tuned with solver-specific values given per simulator below.

    \begin{table}[htbp]
        \centering
        \footnotesize
        \setlength{\tabcolsep}{4pt}
        \caption{\textbf{Simulator backbone configurations} under the shared cloth asset, MPPI controller, action clamp ($20$~mm/step), and cost. Per-solver iteration counts use different update rules and are not directly comparable as a compute proxy.}
        \label{tab:app_exp_simulator_config}
        \begin{tabular}{lcccc}
            \toprule
             & \textsc{Genesis} & \textsc{Newton} & \textsc{Isaac Sim} & \textsc{FLASH} \\
            \midrule
            Dynamic solver          & PBD     & VBD     & XPBD    & FEM \\
            Solver iters         & $4$     & $1$     & $16$    & $8$ \\
            Time step $\Delta t$ & $10$~ms & $10$~ms & $10$~ms & $10$~ms \\
            Friction $\mu$       & $1.0$   & $1.0$   & $1.0$   & $1.0$ \\
            \bottomrule
        \end{tabular}
    \end{table}

    \textbf{\textsc{Genesis}.} \textsc{Genesis} uses position-based dynamics (PBD) with $4/1/1$ iterations on stretch/bending/shear. It re-tessellates the input into $\sim\!1400$ internal particles while keeping $464$ controllable vertices, with stretch compliance $\alpha_{\mathrm{stretch}}=10^{-4}$, bending compliance $\alpha_{\mathrm{bend}}=10^{-2}$, and density $1.0$.

    \textbf{\textsc{Newton}.} \textsc{Newton} uses vertex block descent (VBD) on Warp with $1$ iteration per step. The material is parameterized by triangle stretch coefficient $200$, edge bending coefficient $5\!\times\!10^{-3}$, and density $10$.

    \textbf{\textsc{Isaac Sim}.} \textsc{Isaac Sim} uses the particle-cloth model, wrapping PhysX's XPBD solver with $16$ iterations per step. The material is parameterized by spring stretch stiffness $10^{4}$ and bending stiffness $2\!\times\!10^{-2}$, and per-particle mass $10^{-3}$.

    \textbf{\textsc{FLASH}.} \textsc{FLASH} uses FEM with isometric bending and a Signorini--Coulomb non-smooth Newton contact solver, with $8$ local--global iterations and $10$ constraint iterations. The material is parameterized by stretch stiffness $4\!\times\!10^{6}$, bending stiffness $2\!\times\!10^{-2}$, and total mass $1.0$.

    \textbf{\textsc{Genesis} fails the fold task.}
    \textsc{Genesis} does not produce a usable fold across any $K$ or seed since the plane friction does not take an effect under drag for the held corner to settle on the target. Tuning substeps, compliance, friction, and pin strategy closed only part of the gap, so we omit \textsc{Genesis} from Table~\ref{tab:simulator_eval}.

    \textbf{Results analysis.}
    (i)~\emph{Compute.} \textsc{Newton} is $2.5$--$3.5\times$ slower per step than \textsc{Isaac Sim}. \textsc{Isaac Sim} and \textsc{FLASH} both scale near-linearly in $K$, with \textsc{Isaac Sim} a constant $\sim\!1.7\times$ faster per step.
    (ii)~\emph{Control accuracy.} \textsc{Newton} and \textsc{Isaac Sim} stay at $6$--$13$~cm KP err.\ with SR\,$\le\!60\%$ at every $K$, while \textsc{FLASH} attains $1.5$--$2.1$~cm and $95$--$100\%$ SR. The gap is not closed by more rollouts. 
    (iii)~\emph{Physical fidelity.} \textsc{Newton} and \textsc{Isaac Sim} sit at median MSE $30$--$300$~cm$^{2}$ and EMD $5$--$10$~cm at every $K$, while \textsc{FLASH} is one to two orders tighter (median MSE $3$--$4$~cm$^{2}$, EMD $1.5$--$2.5$~cm). Furthermore, \textsc{FLASH} is the only candidate that provides a usable physics oracle for online MPPI cloth folding across our $K$ range.

\subsection{State Estimation}
\label{app:exp_state_estimation}

    \textbf{Cloth assets.}
    We evaluate on two cloth assets, a towel and a long-sleeve top. The towel uses a sparse triangle mesh of $N_{\mathrm{towel}}=464$ vertices. The long-sleeve top uses $N_{\mathrm{sleeve}}=1379$ vertex sparse mesh. Both meshes and their vertex orderings are computed once and persisted; the same per-asset vertex set is used by the FLASH rollout backend and as the ground-truth labels for state estimation.

    \textbf{Data collection.}
    Episodes are generated in \textsc{FLASH} with scripted goal-guided demonstration trajectories that drive the cloth toward target fold configurations or settle it at rest. Per asset, we collect two subsets. The \emph{fold} subset samples random goal keypoints on a workspace grid so each episode covers a different fold target. The \emph{rest-state} subset drops the cloth under gravity from a randomized initial pose and stores one frame per episode at rest.

    At episode reset, the cloth is randomized by drawing the centroid offset $\Delta x \sim \mathcal{U}(0.16, 0.40)$~m and $\Delta z \sim \mathcal{U}(-0.40, -0.20)$~m, and the yaw angle $\sim \mathcal{U}(-0.50, 0.50)$~rad.

\begin{figure}[htbp]
        \centering
        \includegraphics[width=\linewidth]{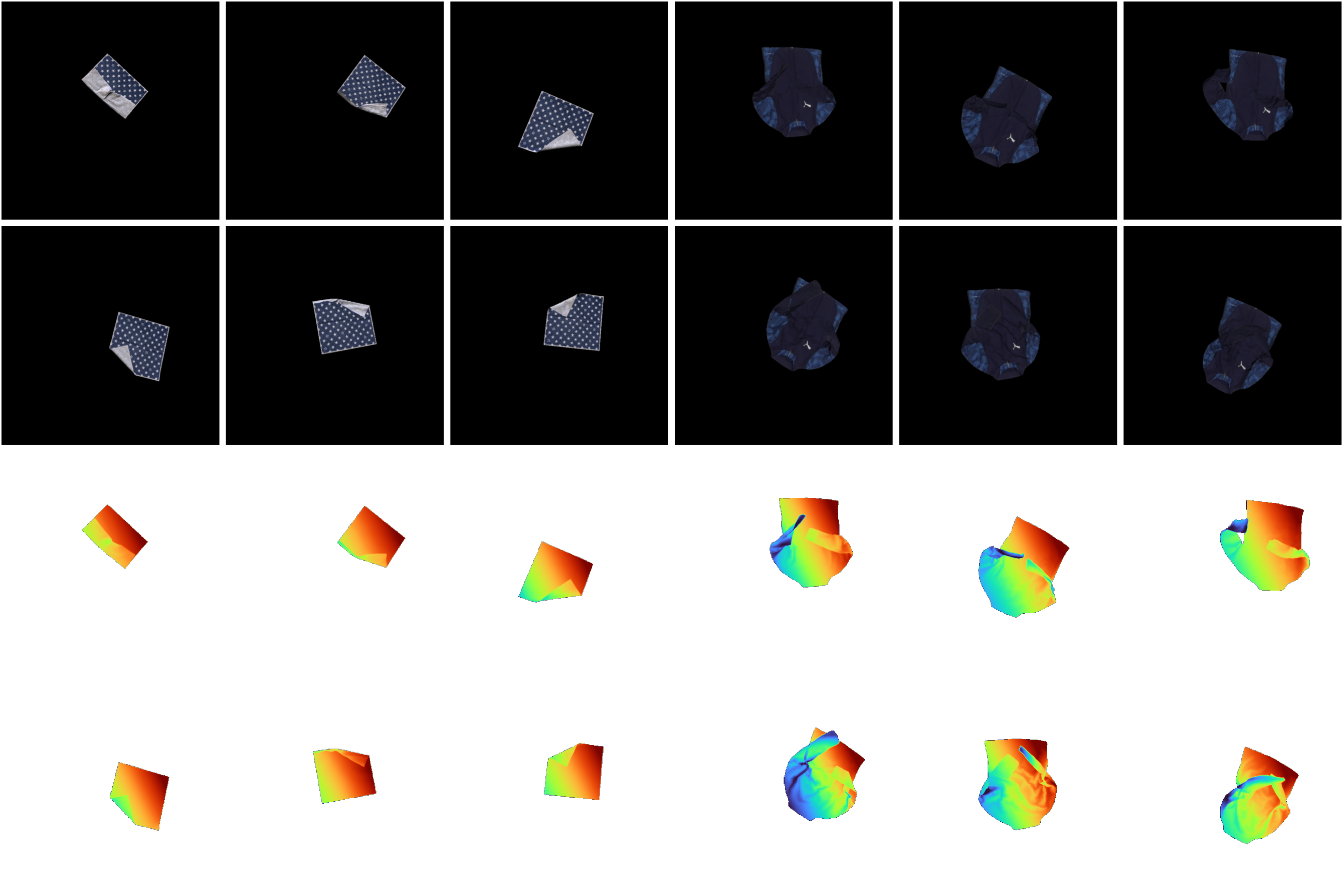}
        \caption{Representative synthetic cloth deformation samples from the state-estimation dataset, covering the towel and the long-sleeve top across the fold and rest-state subsets. Top, RGB views. Bottom, depth maps.}
        \label{fig:app_data_samples}
    \end{figure}

    Each frame logs the RGB image, the depth map, and the ground-truth mesh-vertex positions in the world frame, written as a \texttt{.pt} file per episode. The point cloud is obtained by back-projecting valid depth pixels through the camera intrinsics. Representative samples are shown in Fig~\ref{fig:app_data_samples}.

    \textbf{Baseline configurations.}
    All three methods predict the same $N$ indexed material vertices per asset and are trained on the dataset described above. They differ in input modality and encoder. Table~\ref{tab:app_exp_state_baselines} summarizes the architecture and training, with per-method notes below.

    \begin{table}[htbp]
        \centering
        \footnotesize
        \setlength{\tabcolsep}{4pt}
        \caption{Architecture and training configuration of the three state-estimation methods.}
        \label{tab:app_exp_state_baselines}
        \resizebox{\linewidth}{!}{%
        \begin{tabular}{lccc}
            \toprule
             & Ours (RGB) & DeFM (depth) & UniClothDiff DPM (point cloud) \\
            \midrule
            Input             & RGB $392^2$                  & depth $392^2$                  & $10{,}000$ points \\
            Encoder           & DINOv2 ViT-S/14, frozen      & DeFM ViT-S/14, frozen          & KNN-grouper $+$ patch MLP \\
            Output head       & \makecell[c]{$N$ canon.\ tokens, \\ $4$L self-attn ($d{=}256$, $h{=}8$), MLP} & same as Ours & \makecell[c]{DDPM denoiser, \\ template-cond.} \\
            Loss              & smooth $\ell_1$              & smooth $\ell_1$                & DDPM noise pred. \\
            Optimizer (AdamW) & lr $3\mathrm{e}{-4}$, wd $10^{-4}$ & lr $3\mathrm{e}{-4}$, wd $10^{-4}$ & lr $10^{-3}$, wd $10^{-2}$ \\
            Batch size        & $32$                         & $32$                           & $4$ per GPU \\
            Schedule          & \makecell[c]{$50$ ep., cosine, \\ $3$-ep.\ warmup, EMA $0.999$} & same as Ours & \makecell[c]{$900$K steps, const., \\ $500$-step warmup} \\
            Gradient clip     & $1.0$                        & $1.0$                          & $1.0$ \\
            \bottomrule
        \end{tabular}%
        }
    \end{table}

\subsection{Real-World Evaluation}
\label{app:exp_realworld}

    We deploy on the AIRBOT Play robot arm in single-arm and dual-arm configurations. Visual observations come from a single ZED 2i stereo camera mounted overhead. Our pipeline uses only the left RGB stream, and the stereo depth is used to provide the point-cloud input required by one baseline. All policy distillation and online planning run on a single NVIDIA RTX 4090. 

    \textbf{Robot Platform.}
    \label{app:hardware}
    We use the AIRBOT Play 6-DoF robot arm in both single-arm and dual-arm configurations. Each arm is commanded through end-effector delta-position. The gripper is a parallel gripper with a binary open and close command. In the dual-arm configuration, the two arms are mounted symmetrically on a common tabletop base. End-effector position commands for both arms are synchronized through a single control loop.

    \textbf{Camera Setup.}
    \label{app:zed}
    Visual observations come from a single ZED~2i stereo camera mounted overhead to provide a good view of the manipulation region. Only the left RGB stream is consumed by the real-to-sim module. The camera runs at resolution $1280 \times 720$.

    \textbf{Success Accounting and Grasp Success Rate.}
    Every trial reported in the main paper is scored end to end from a randomized initial cloth pose. A trial counts as a success only if the commanded grasp acquires the cloth \emph{and} the final corner alignment meets the task criterion. A failed grasp is counted as a failed trial. Reported separately, every required grasp succeeds in $10/10$ single-arm diagonal trials and in $9/10$ dual-arm symmetric trials of the full-pipeline rows of Tab.~\ref{tab:realworld}, against end-to-end success rates of $9/10$ and $8/10$. In the mid-fold disturbance setting of Tab.~\ref{tab:generalization} the grasp succeeds in $10/10$ trials and the fold is completed in $8/10$. The dual-arm task requires two simultaneous grasps, so a single missed corner costs the trial.

    \textbf{Per-Step Latency.}
    On a single RTX~4090, the computation in one control step is $95.8$~ms of segmentation, $8.5$~ms of real-to-sim inference, $0.5$~ms of state writing into the $K$ resident environments, and $970$~ms of MPPI at $H\!=\!5$ on the $464$-vertex towel. On the $1156$-vertex shorts the MPPI term rises to $2550$~ms while real-to-sim inference rises only to $11.6$~ms. Rollout throughput, not perception, is therefore the binding constraint, and it scales with mesh resolution.

\begin{figure}[t]
    \centering
    \includegraphics[width=\linewidth]{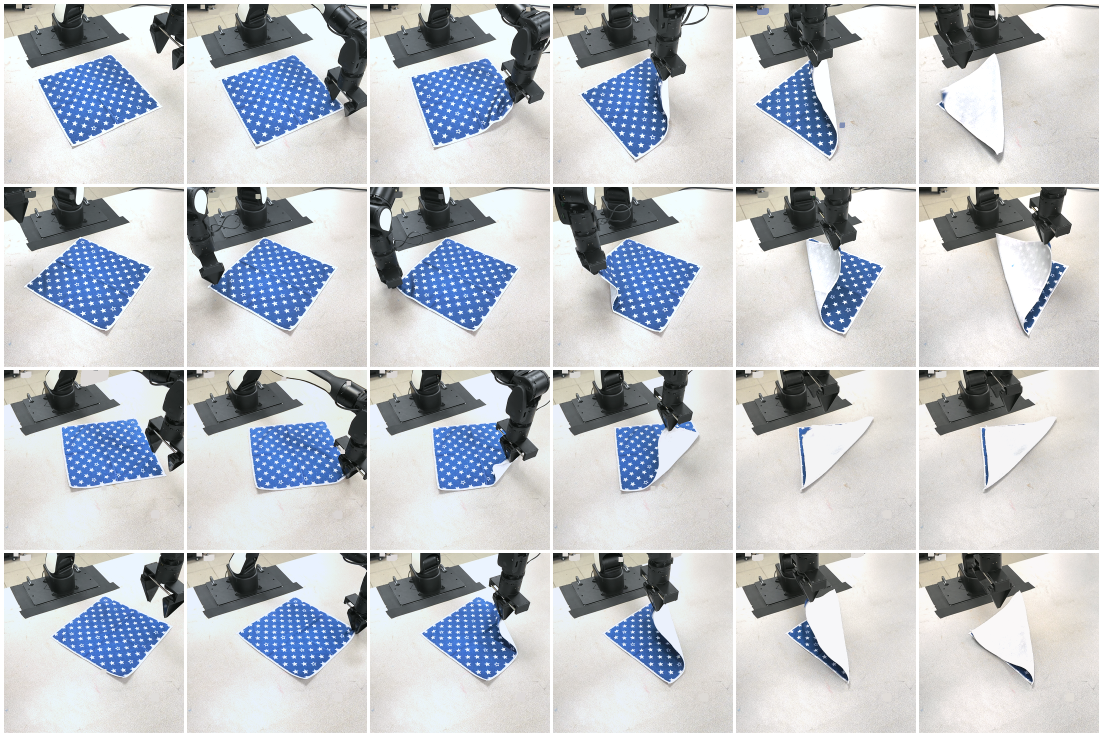}
    \caption{
    Full-rollout visualization of successful single-arm diagonal folding trials. Time progresses from left to right.
    }
    \label{fig:app_realworld_rollout_1}
\end{figure}
\begin{figure}[t]
    \centering
    \includegraphics[width=\linewidth]{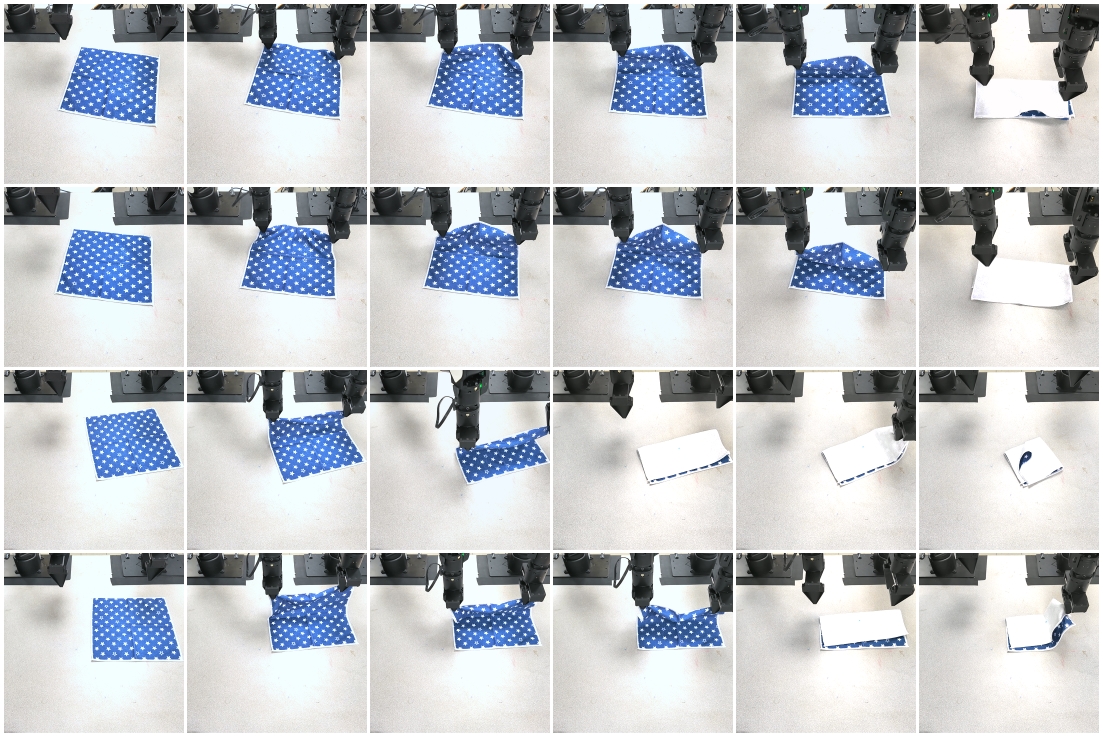}
    \caption{
    Full-rollout visualization of successful dual-arm symmetric folding trials. Time progresses from left to right.
    }
    \label{fig:app_realworld_rollout_2}
\end{figure}

\paragraph{Visualizations.}
Additional full-rollout visualizations of successful real-world trials are provided in Fig.~\ref{fig:app_realworld_rollout_1}--Fig.\ref{fig:app_realworld_rollout_2}.
    
\subsection{Generalization and Closed-Loop Correction}
\label{app:exp_generalization}

This section provides detailed configurations for the four generalization and disturbance-recovery settings, together with additional trial visualizations. Each trial randomizes the initial garment pose by drawing a translation uniformly from $[-5,\, 5]$~cm along the two horizontal axes and a planar rotation from $[-0.5,\, 0.5]$~rad.

\begin{figure}[t]
    \centering
    \includegraphics[width=\linewidth]{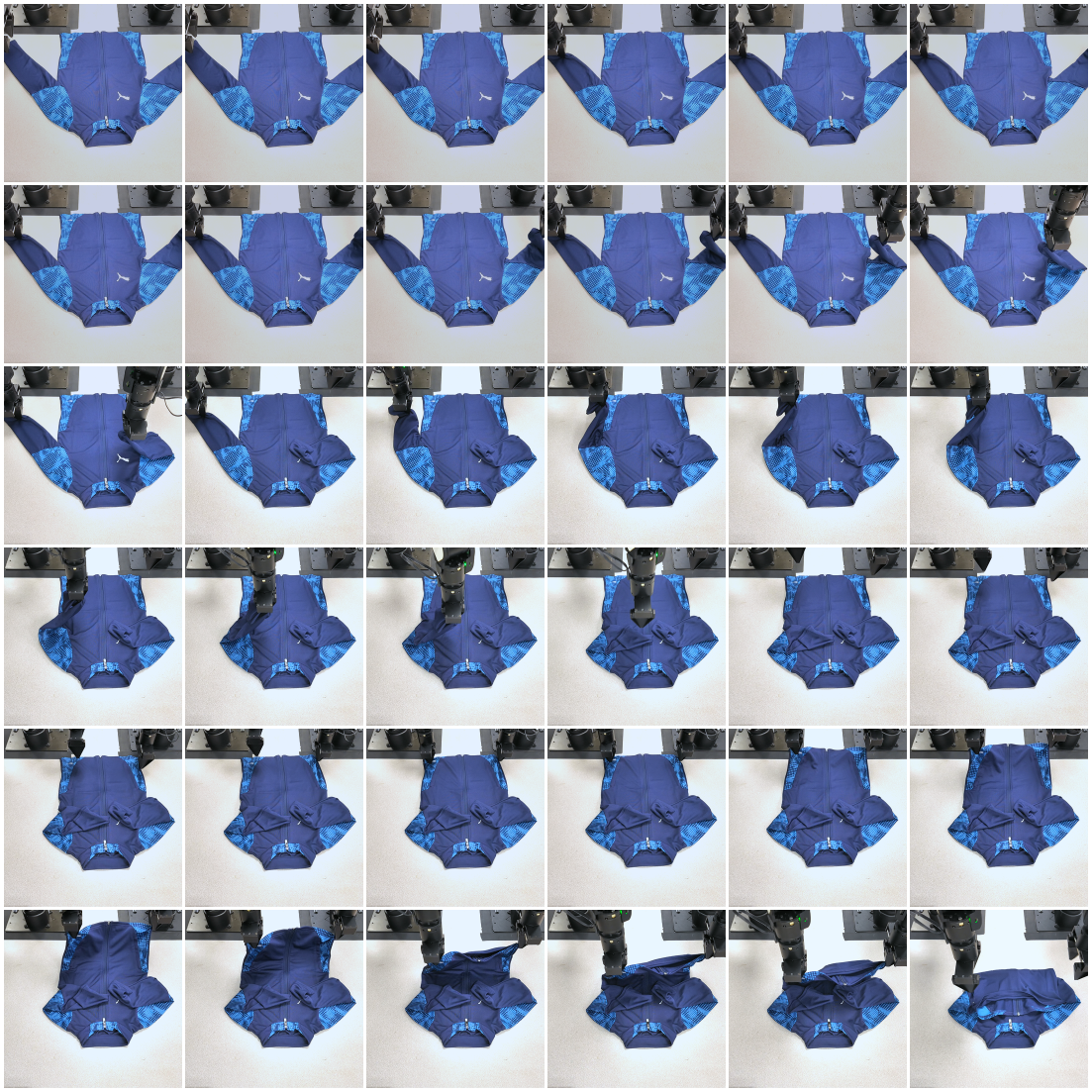}
    \caption{Full rollout visualization for long-sleeve shirt folding.}
    \label{fig:app_longsleeve_rollout}
\end{figure}

\textbf{Long-sleeve shirt folding.}
The task folds both sleeves toward the torso center and then folds the torso along the horizontal midline. Full rollouts are visualized in Fig.~\ref{fig:app_longsleeve_rollout}.

\begin{figure}[t]
    \centering
    \includegraphics[width=\linewidth]{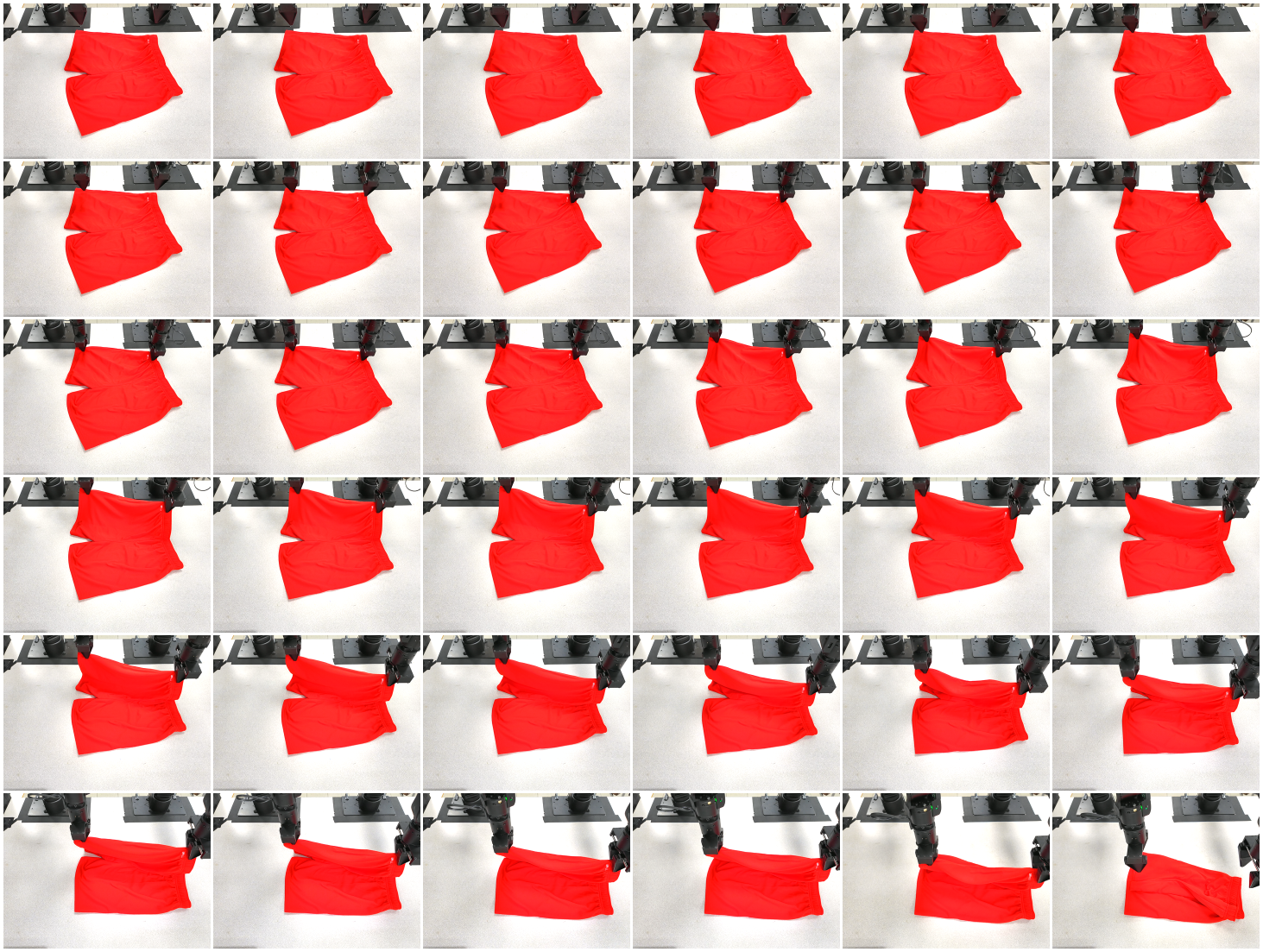}
    \caption{Full rollout visualization for shorts folding.}
    \label{fig:app_shorts_rollout}
\end{figure}

\textbf{Shorts folding.}
The task folds the two legs toward the waistband region and then folds the shorts into a compact final shape. Full rollouts are visualized in Fig.~\ref{fig:app_shorts_rollout}.

\begin{figure}[t]
    \centering
    \includegraphics[width=\linewidth]{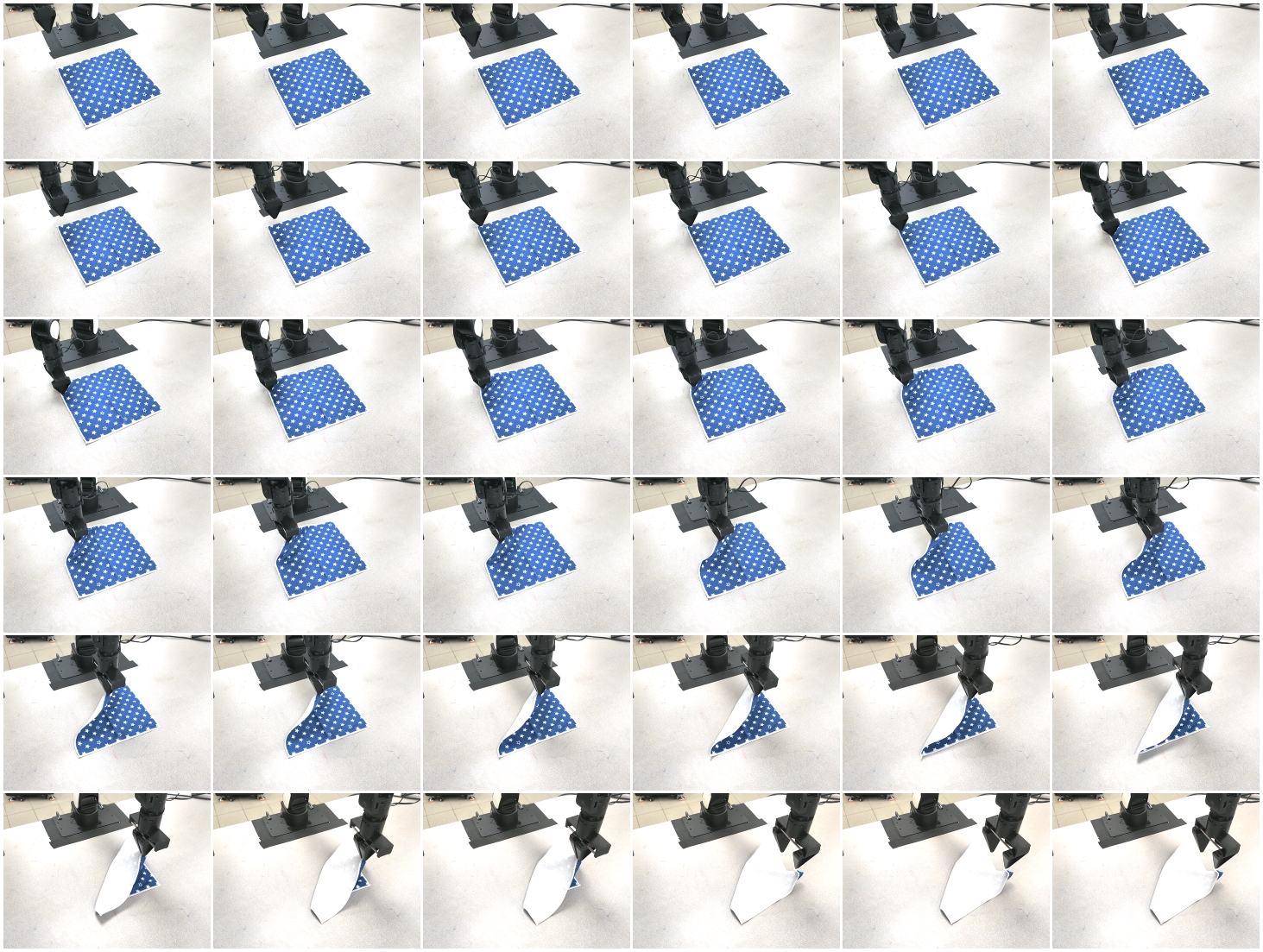}
    \caption{Full rollout visualization for reverse-diagonal towel folding.}
    \label{fig:app_reverse_diagonal_rollout}
\end{figure}

\textbf{Reverse-diagonal towel folding.}
The task folds the same square towel along the opposite diagonal from the one used in the main single-arm experiment, without retraining. Full rollouts are visualized in Fig.~\ref{fig:app_reverse_diagonal_rollout}.

\begin{figure}[t]
    \centering
    \includegraphics[width=\linewidth]{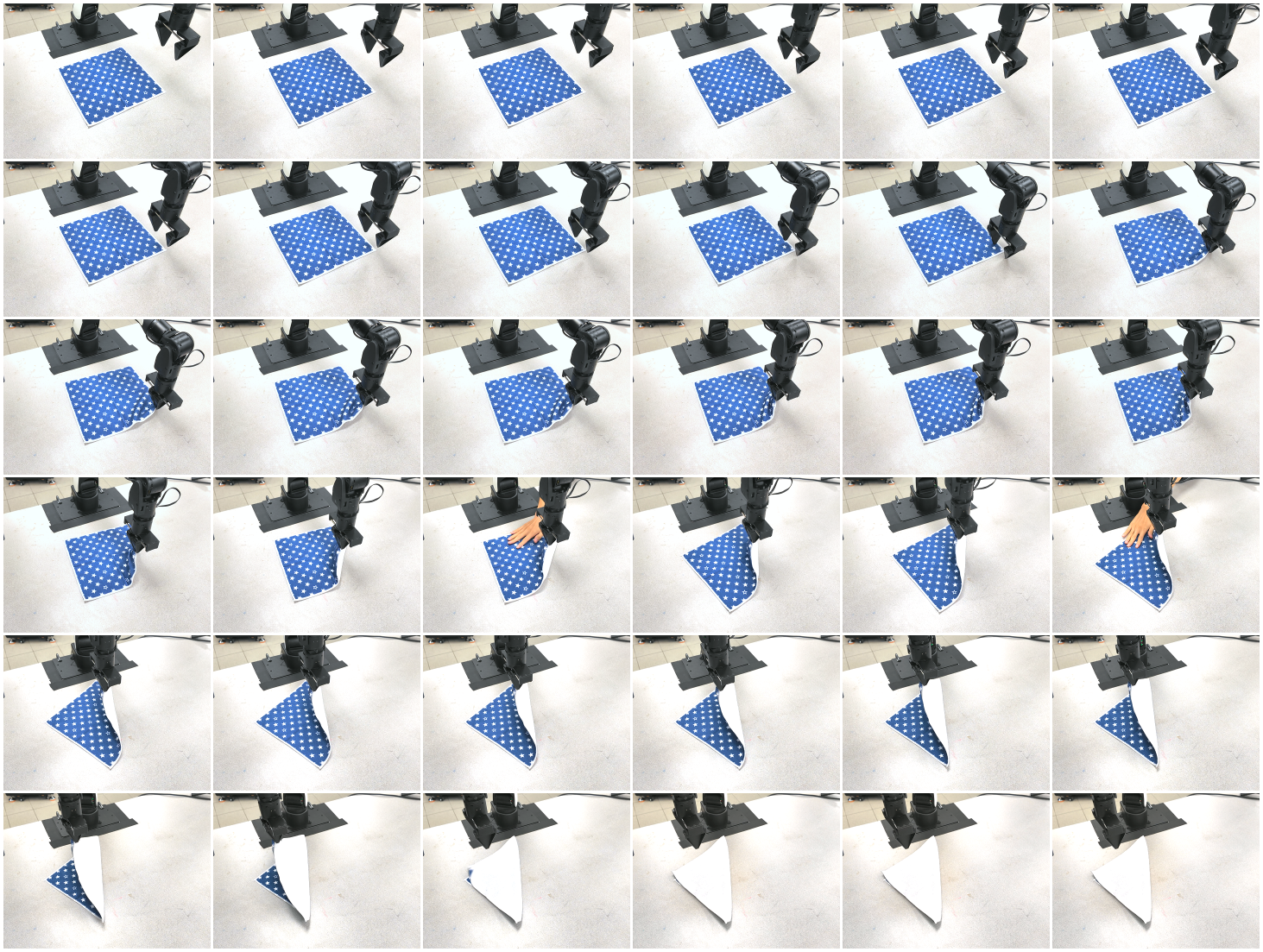}
    \caption{Full rollout visualization for mid-fold disturbance recovery.}
    \label{fig:app_disturbance_rollout}
\end{figure}

\textbf{Mid-fold disturbance recovery.}
The task tests whether the system can recover when the towel is manually displaced during the folding process and continue folding from the new state. Full rollouts are visualized in Fig.~\ref{fig:app_disturbance_rollout}.

\textbf{What varies across the four settings.} Relative to the towel training distribution, geometry and topology change from the $464$-vertex towel to the notched $1156$-vertex shorts and the $1379$-vertex long-sleeve top, and mass, bending, and friction are tuned per garment accordingly. The fold sequence changes from a single stage to two stages, and the arm count from one to two. The reverse-diagonal setting holds the asset fixed and instead swaps the pick and goal keypoints relative to those the base policy was distilled on. Appearance variation is covered by the training-time randomization of Tab.~\ref{tab:app_rgb_aug}.

\clearpage
\section{Additional Experimental Analyses}
\label{app:reb_details}

\begin{figure}[htbp]
    \centering
    \begin{subfigure}{0.48\linewidth}
        \centering
        \includegraphics[width=\linewidth]{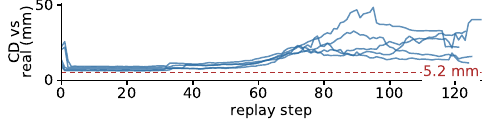}
        \caption{Open-loop simulator replay vs. real cloth}
        \label{fig:replay_error}
    \end{subfigure}
    \hfill
    \begin{subfigure}{0.48\linewidth}
        \centering
        \includegraphics[width=\linewidth]{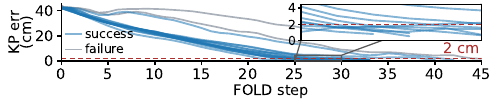}
        \caption{Closed-loop keypoint error}
        \label{fig:closed_loop_error}
    \end{subfigure}
    \caption{\textbf{Simulator fidelity and closed-loop correction.}
    (a) One-sided Chamfer distance between the \textsc{FLASH} mesh and the real cloth when \emph{real} single-arm end-effector trajectories are replayed open-loop, one line per trajectory.
    (b) Per-step keypoint error of the closed-loop planner in simulation, one line per trial over $20$ seeds.}
    \label{fig:reb_quant}
\end{figure}

\subsection{Simulator Fidelity Against Real Cloth Trajectories}
\label{app:reb_fidelity}
Inter-simulator comparison establishes relative accuracy but not fidelity to real fabric. We therefore score \textsc{FLASH} against real recordings following the one-sided Chamfer protocol of Blanco-Mulero et al.~\cite{blancomulero2024benchmarking}, which measures, for each simulated mesh vertex, the distance to the nearest point of the recorded cloth point cloud.

\textbf{Protocol.} A single-arm diagonal fold is executed on the robot and recorded with the overhead ZED~2i in RGB-D. We replay the end-effector trajectory of each real execution through \textsc{FLASH}, initializing the cloth once from the first frame and never re-synchronizing afterwards, so what is reported is open-loop divergence rather than the tracking error of the closed-loop system. The evaluation uses $5$ trajectories and $626$ frames spanning approach, grasp, lift, drag and release.

\textbf{Results.} The simulated mesh tracks the real cloth to a median of $13.6$~mm, and to $7$--$9$~mm through the approach and grasp phases, with error growing during drag as the open-loop replay accumulates divergence (Fig.~\ref{fig:replay_error}). This figure includes a measurement floor. Registering the mesh to frames in which the cloth is flat and undisturbed already costs $5.2$~mm under this metric, reflecting camera calibration, segmentation boundary and mesh discretization rather than simulator error.

\subsection{State-Estimation Error on Real Observations}
\label{app:reb_estimator}
The comparison in Tab.~\ref{tab:state_estimation} scores the estimator against simulator ground truth. To quantify the sim-to-real gap directly, we additionally record depth alongside RGB and score the predicted cloth state against the back-projected real point cloud using the same one-sided Chamfer distance, on the same $5$ trajectories and $626$ frames. The estimator reaches $10.2\pm1.4$~mm per trajectory. This is below the $13.6$~mm of the open-loop replay because the estimator re-observes the cloth at every control step rather than extrapolating the fold forward from a single initialization. The estimator itself never receives depth, which is recorded here only for scoring.

\subsection{Closed-Loop Error Trajectories and Disturbance Magnitude}
\label{app:reb_closedloop}
We measure the correction in simulation over $20$ seeds of the single-arm diagonal fold under the deployed prior-guided controller, with randomized initial pose.

\textbf{Error trajectories.} Fig.~\ref{fig:closed_loop_error} plots the planner's keypoint error at every control step. The error measured here is the distance between the picked keypoint and its target, which starts at the $42.2 \pm 0.2$~cm diagonal of the $30$~cm towel, that is, the full distance the grasped corner has to travel. Eighteen of the $20$ seeds reach the $2$~cm success criterion, ending at $0.8 \pm 0.1$~cm after $22$ to $31$ control steps. The trajectories consistently decrease toward the success criterion rather than plateauing above it, providing direct evidence of progressive closed-loop error correction.

\textbf{Disturbance magnitude.} Eight control steps into the fold, the cloth is pushed by a fixed displacement in a direction drawn per seed, with $20$ seeds at each magnitude. The keypoint error returns below its pre-disturbance level within $4$ to $6$ control steps at every magnitude (Tab.~\ref{tab:app_disturbance}).

    \begin{table}[htbp]
        \centering
        \footnotesize
        \setlength{\tabcolsep}{8pt}
        \caption{\textbf{Mid-fold disturbance sweep in simulation.} A push of the stated magnitude is applied eight control steps into the fold, $20$ seeds per magnitude.}
        \label{tab:app_disturbance}
        \begin{tabular}{ccc}
            \toprule
            Push (cm) & Seeds & Recovered within (steps) \\
            \midrule
            $6$  & $20$ & $4.0 \pm 0.0$ \\
            $9$  & $20$ & $4.1 \pm 0.7$ \\
            $12$ & $20$ & $6.0 \pm 1.0$ \\
            \bottomrule
        \end{tabular}
    \end{table}

\subsection{Reproduction of the DPM Baseline}
\label{app:reb_dpm}
The absolute CD we report for DPM in Tab.~\ref{tab:state_estimation} is not comparable to the value reported in~\cite{pmlr-v305-tian25c}, because the evaluation setting and input configuration differ. We retrain DPM from scratch on our garments, our mesh topology ($N\!=\!464$ for the towel and $N\!=\!1379$ for the long-sleeve top) and our fold distribution. We give it a single overhead cloud rather than its original multi-view input, and keep the original $50$ denoising steps. Raising this to $1000$ changes the metrics by under $5\%$ at twenty times the inference cost.

\section{Real-to-Sim Augmentation Ablation}
\label{app:ablation_r2s_aug}
This section ablates the effect of training-time data augmentation for the RGB-native real-to-sim module. We train the same model with and without the input- and latent-level randomization, and evaluate reconstruction accuracy and real-world closed-loop success.
    \begin{table}[htbp]
        \centering
        \caption{\textbf{Randomization ablation.} Validation reconstruction metrics and real-world success rate over $10$ trials.}
        \label{tab:e4_rgb_aug}
        \small
        \setlength{\tabcolsep}{5pt}
        \begin{tabular}{llcccc}
          \toprule
          Task & R2S Aug & MSE (mm$^2$) & CD (mm) & EMD (mm) & Real SR ($\uparrow$) \\
          \midrule
          \multirow{2}{*}{\makecell[l]{Single-arm \\ diagonal}}
            & on (Ours) & \cellcolor{ourblue}2.33$\pm$1.21 & \cellcolor{ourblue}1.94$\pm$0.41 & \cellcolor{ourblue}1.98$\pm$0.45 & \cellcolor{ourblue}\textbf{9/10} \\
            & off       & 1.85$\pm$1.93 & 1.72$\pm$0.47 & 1.74$\pm$0.51 & 6/10 \\
          \midrule
          \multirow{2}{*}{\makecell[l]{Dual-arm \\ symmetric}}
            & on (Ours) & \cellcolor{ourblue}3.59$\pm$2.08 & \cellcolor{ourblue}2.42$\pm$0.57 & \cellcolor{ourblue}2.51$\pm$0.68 & \cellcolor{ourblue}\textbf{8/10} \\
            & off       & 3.64$\pm$2.05 & 2.37$\pm$0.58 & 2.42$\pm$0.60 & 5/10 \\
          \bottomrule
        \end{tabular}
    \end{table}
Table~\ref{tab:e4_rgb_aug} shows that training-time augmentation has little effect on reconstruction accuracy on the synthetic test set, but becomes important during real-world deployment. Without augmentation, the estimator is less robust to real-image noise, visual-domain shift, and robot-arm or gripper occlusions, leading to poorer state synchronization and lower closed-loop success.

\end{document}